%% file: main.tex
\documentclass[10pt,twocolumn,letterpaper]{article}

\usepackage[pagenumbers]{cvpr} 

\usepackage{graphicx}
\usepackage{amsmath}
\usepackage{amssymb}
\usepackage{booktabs}
\usepackage{float}
\usepackage{mathtools}
\usepackage{stfloats}
\usepackage{tabularx}
\usepackage{adjustbox}
\usepackage{colortbl}
\usepackage{graphicx}
\usepackage{amsmath}
\usepackage{amssymb}
\usepackage{booktabs}
\usepackage{float}
\usepackage{mathtools}
\usepackage{stfloats}
\usepackage{tabularx}
\usepackage{adjustbox}
\usepackage{colortbl}
\usepackage{algorithmic,algorithm}
\usepackage{enumitem}
\usepackage[accsupp]{axessibility}
\newcommand\blfootnote[1]{%
  \begingroup
  \renewcommand\thefootnote{}\footnote{#1}%
  \addtocounter{footnote}{-1}%
  \endgroup
}
\usepackage[pagebackref,breaklinks,colorlinks]{hyperref}

\usepackage[capitalize]{cleveref}
\crefname{section}{Sec.}{Secs.}
\Crefname{section}{Section}{Sections}
\Crefname{table}{Table}{Tables}
\crefname{table}{Tab.}{Tabs.}

\def\cvprPaperID{155} 
\def\confName{CVPR}
\def\confYear{2022}

\begin{document}

\title{Maximum Spatial Perturbation Consistency for Unpaired Image-to-Image Translation}

\author{
  {Yanwu Xu$^1$, Shaoan Xie$^2$, Wenhao Wu$^4$, Kun Zhang$^2$, Mingming Gong$^3$\footnote[2]{}, Kayhan Batmanghelich$^1$\footnote[2]{}}\\
  \tt\small $^1$Department of Biomedical Informatics, University of Pittsburgh, \\ \tt\small \{yanwuxu,kayhan\}@pitt.edu \\
  \tt\small $^2$Department of Philosophy, Carnegie Mellon University \\
  \tt\small $^3$School of Mathematics and Statistics, The University of Melbourne \\
  \tt\small $^4$Department of Computer Vision Technology (VIS), Baidu Inc.
}
\maketitle

\input{abstract_v2}

\input{introduction_v3}
\input{related_work}
\input{method_v1}

\input{experiment}

\input{conclusion}
{\small
\bibliographystyle{ieee_fullname}
\bibliography{main.bbl}
}

\input{Supplementary}

\end{document}

%% file: abstract_v2.tex
\begin{abstract}
\vspace{-4mm}
Unpaired image-to-image translation (I2I) is an ill-posed problem, as an infinite number of translation functions can map the source domain distribution to the target distribution. Therefore, much effort has been put into designing suitable constraints, e.g., cycle consistency (CycleGAN), geometry consistency (GCGAN), and contrastive learning-based constraints (CUTGAN), that help better pose the problem. 
However, these well-known constraints have limitations: (1) they are either too restrictive or too weak for specific I2I tasks; (2) these methods result in content distortion when there is a significant spatial variation between the source and target domains.
This paper proposes a universal regularization technique called maximum spatial perturbation consistency (MSPC), which enforces a spatial perturbation function ($T$) and the translation operator ($G$) to be commutative (i.e., $T \circ G = G \circ T $). In addition, we introduce two adversarial training components for learning the spatial perturbation function. The first one lets $T$ compete with $G$ to achieve maximum perturbation. The second one lets $G$ and $T$ compete with discriminators to align the spatial variations caused by the change of object size, object distortion, background interruptions, etc. 
Our method outperforms the state-of-the-art methods on most I2I benchmarks. We also introduce a new benchmark, namely the front face to profile face dataset, to emphasize the underlying challenges of I2I for real-world applications. We finally perform ablation experiments to study the sensitivity of our method to the severity of spatial perturbation and its effectiveness for distribution alignment.
\blfootnote{ $\dagger$ Equal advising. Code and the new face data is released at \url{https://github.com/batmanlab/MSPC}.}
\end{abstract}

%% file: introduction_v3.tex
\section{Introduction}

In unpaired image-to-to image translation (I2I), one aims to translate images from a source domain $\mathcal{X}$ to a target domain $\mathcal{Y}$, with data drawn from the marginal distribution of the source domain ($P_X$) and that of the target domain ($P_Y$). Unpaired I2I has many applications, such as super-resolution \cite{superresolution,superresolution2}, image editing \cite{imageediting,imageediting2}, and image denoising \cite{imagedenoising,imagedenoising2}. However, it is an ill-posed problem, as there is an infinite choice of translators $G$ that can map $P_X$ to $P_Y$. 
\begin{figure}[t]
\centering
\subfloat[Consistency regularization with spatial perturbation function $T$]{
	\label{intro_perturbation}
	\includegraphics[width=0.25\textwidth]{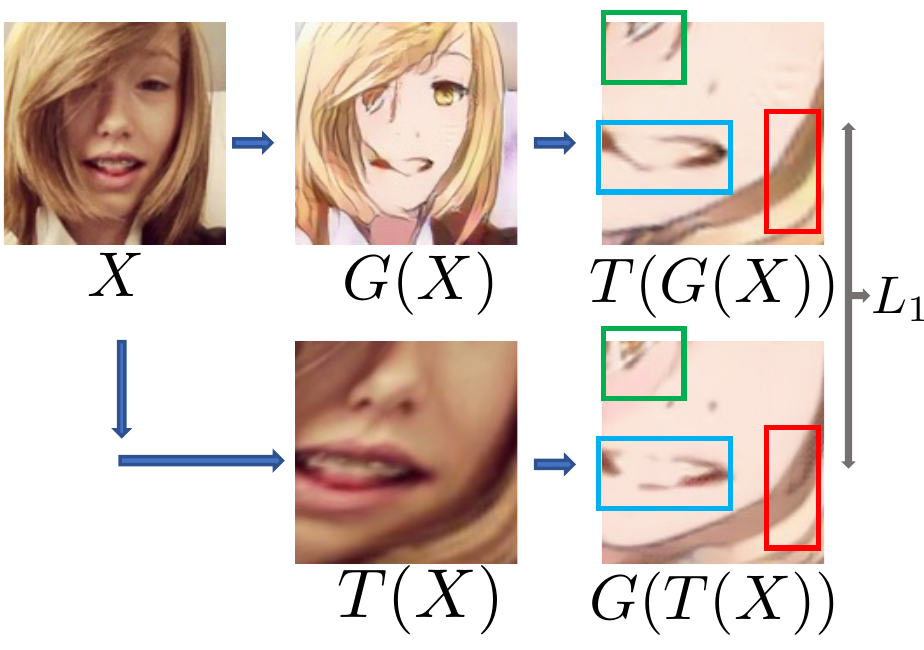} } 

\subfloat[Spatial alignment of spatial perturbation function $T$]{
	\label{intro_alignment}
	\includegraphics[width=0.25\textwidth]{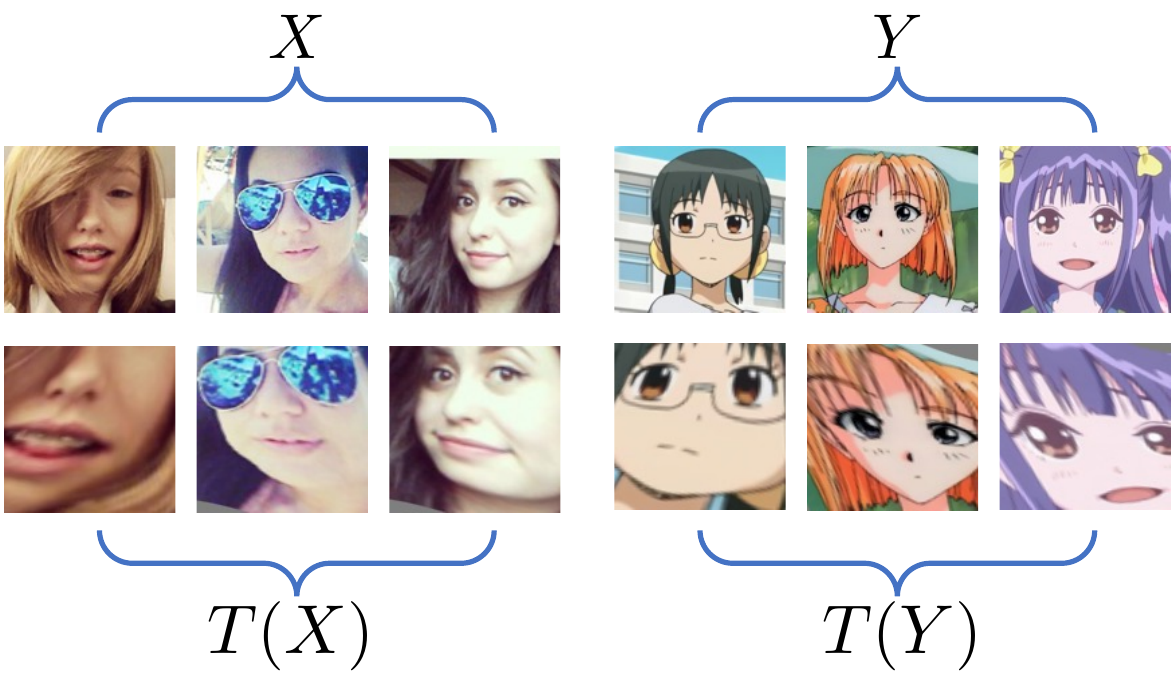} } 
\caption{In this figure, we illustrate the the proposed MSPC on (a) consistency regularization under maximum spatial perturbation and (b) aligning the spatial distributions between source $X_T$ and $Y_T$ via spatial perturbation function $T$.}
\label{intro}
\end{figure}

Various constraints on the translation function $G$ have been proposed to remedy the ill-posedness of the problem. For example, cycle consistency (CycleGAN) \cite{cycle} enforces the cyclic reconstruction consistency: $X\to G(X) \to X$, which means $G$ and its inverse are bijections. CUTGAN \cite{cut} maximizes the mutual information between an input image and the translated image via constrastive learning on the patch-level features.
The GCGAN~\cite{gcgan}, on the other hand, effectively uses geometric consistency by applying a predefined geometry transformation $g$, i.e., fixed rotation, encouraging $G$ to be robust to geometry transformation. The underlying assumption of the GCGAN is that the $G$ and $g$ are commutative (i.e., $g \circ G = G \circ g $).
However, CycleGAN assumes that the relationship of bijection between source and target, which is limited for most real-life applications \cite{cut}. For instance, the translation function is non-invertible in the \emph{Cityscapes} $\to$ \emph{Parsing} task. Though geometry consistency used in GCGAN is a general I2I constraint, it is too weak in the sense that the model would easily memorize the pattern of a fixed transformation. 
CUTGAN enforces the strong correlation between the input images and the translated images at the corresponded patches; thus it would fail when the patches at the same spatial location do not contain the same content, \eg, in the \emph{Front Face} $\to$ \emph{Profile} task (shown in Figure \ref{face failure}). Thus, the above models are either too restrictive or too weak for specific I2I tasks. Besides, all of them overlook the extra spatial variations in image translation, which are caused by the change of object size, object distortion, background interruptions, \etc.

To tackle the issues above, we propose a novel regularization called the maximum spatial perturbation consistency (MSPC), which enforces a new type of constraint and aligns the content's spatial distribution content across domains. Our MSPC generalizes GCGAN by learning a spatial perturbation function $T$, which adaptively transforms each image with an image-dependent spatial perturbation. Moreover, MSPC is based on the new insight that consistency on hard spatial perturbation would boost the robustness of translator $G$. Thus, MSPC enforces the maximum spatial perturbation function ($T$) and the translation operator ($G$) to be commutative (i.e., $T \circ G = G \circ T $). To generate the maximum spatial perturbation, we introduce a differentiable spatial transformer $T$ \cite{stn} to compete with the translation network $G$ in a mini-max game, which we mark as the perturbation branch. More specifically, $T$ tries to maximize the distance between $T(G(X))$ and $G(T(X))$, and $G$ minimizes the difference between them. In this way, our method dynamically generates the hardest spatial transformation for each image, avoiding overfitting $G$ to specific spatial transformations. The Figure~\ref{intro_perturbation} give a simple illustration of how the image-dependent spatial perturbation works on the I2I framework.


 To align the spatial distribution of the content, $T$ and $G$ cooperate to compete with a discriminator $D_{pert}$ in another mini-max game, which we mark as an alignment branch. In the alignment branch, $T$ participates in aligning the distribution between the translated images and the target images by alleviating the spatial discrepancy, \ie adjusting the object's size, cropping out the noisy background, and further reducing undesired distortions in the translation network $G$. We evaluate our model on several widely studied benchmarks, and additionally, we construct a \emph{Front Face} $\to$ \emph{Profile} dataset with significant domain gaps to emphasize the challenges in real-world applications. The experimental results show that the proposed MSPC outperforms its competitors on most I2I tasks. More importantly, MSPC performs the most stable across various I2I tasks, demonstrating the universality of our constraint. The Figure~\ref{intro_alignment} shows the visual examples the alignment effect on source and target images via dynamic spatial transformation function.

%% file: related_work.tex
\section{Related Work}
\subsection{Generative Adversarial Network}
Generative adversarial networks (GANs)\cite{gan} train a min-max game between the generator $G$ and the discriminator $D$, where $D$ tries to discriminate between the data distribution and the generated distribution. When $G$ and $D$ reach a equilibrium, the generated distribution will exactly match the data distribution. In recent years, GANs have been explored in many image synthesis tasks, such as supervised and unsupervised image generation \cite{sn,projection,bigan,biggan,tac}, domain adaptation \cite{da1,da2,da3}, image inpainting \cite{paint1,paint2,paint3}, \etc.

\subsection{Image-to-Image Translation}

The paired image-to-image translation task can be traced back to \cite{imagetranslation1}, which proposes a non-parametric texture model. With the development of deep learning, the recent Pix2Pix model \cite{googlemap} expands the conditional GAN model to the image translation and learns a conditional mapping from source images to the target images with paired data. There are also other works in this line of research, such as \cite{imagetranslation2,imagetranslation3}. However,  paired images are expensive to collect,  and thus the latest works focus on the setting with semi-supervised and unsupervised settings. Compared to existing unpaired setting,  \cite{unaligned_i2i} considers a more challenging setting where contents of two domains are unaligned and proposes to address this issue with importance re-weighting. As a semi-supervised method, \cite{semi_imagetranslation} performs  image translation with the combined limited paired images and sufficient unpaired images. Furthermore, \cite{cycle,gcgan,cut,unit, choi2018stargan, choi2020stargan, kim2019u, liu2019few, mejjati2018unsupervised, baek2021rethinking, chen2020reusing, yu2019multi, cho2019image, negcut, lsesim} focus on the unsupervised image translation tasks. In these works, CycleGAN \cite{cycle} proposes a cycle consistency between the input images and the translated images. GCGAN \cite{gcgan} minimizes the error translated images via the rotation on the input images. CUTGAN \cite{cut} maximizes the mutual information between the input and the translated images via contrastive learning. UNIT \cite{unit} proposes a strong assumption of content sharing and style change between two image domains in the latent space. To obtain diverse translation results, MUNIT \cite{munit} and DRIT \cite{drit} disentangle the content and the style and generate diverse outputs by combing the same content with different styles. In this paper, we focus on the unsupervised task with deterministic output of image translation.

\subsection{Consistency Regularization of Semi-Supervised Learning}
Among various methods for the semi-supervised classification, clustering, or regression task, consistency regularization has attracted much attention, as discussed in a recent survey paper on deep semi-supervised learning  \cite{semi_survey}. The constraint of consistency regularization assumes that the manifold of data is smooth and that the model is robust to the realistic perturbation on the data points. In other words, consistency regularization can force the model to learn a smooth manifold via incorporating the unlabeled data. Though GCGAN was proposed from a different perspective, it can be considered as a variation of $\Pi$ model \cite{pi}, which enforces consistent model prediction on two random augmentations on a labeled or unlabeled sample.

The regularization method closely related to the proposed MSPC is virtual adversarial training (VAT) \cite{vat}. VAT introduced the concept of adversarial attack \cite{adversarial_attack} as a consistency regularization in  semi-supervised classification. This method learns a maximum adversarial perturbation as a additive noise on the data-level. To be more specific, it finds an optimal perturbation $\gamma$ on a input sample $x$ under the constraint of $\gamma < \delta$. Letting $\mathcal{R}$ and $f$ denote the estimation of distance between two vectors and the predicted model respectively, we can formulate it as:
	\begin{equation}
		\min_f\max_{\gamma;\|\gamma\|\leqslant \delta}\mathbb{E}_{x\in P_X}\mathcal{R}( f( \theta ,x) ,f( \theta ,x+\gamma )).
		\label{equ:VAT}
	\end{equation}

%% file: method_v1.tex
\section{Proposed Method}\label{method}
\begin{figure*}[t]
\centering
\subfloat[Complete Model of MSPC.]{
	\label{model}
	\includegraphics[width=0.6\textwidth]{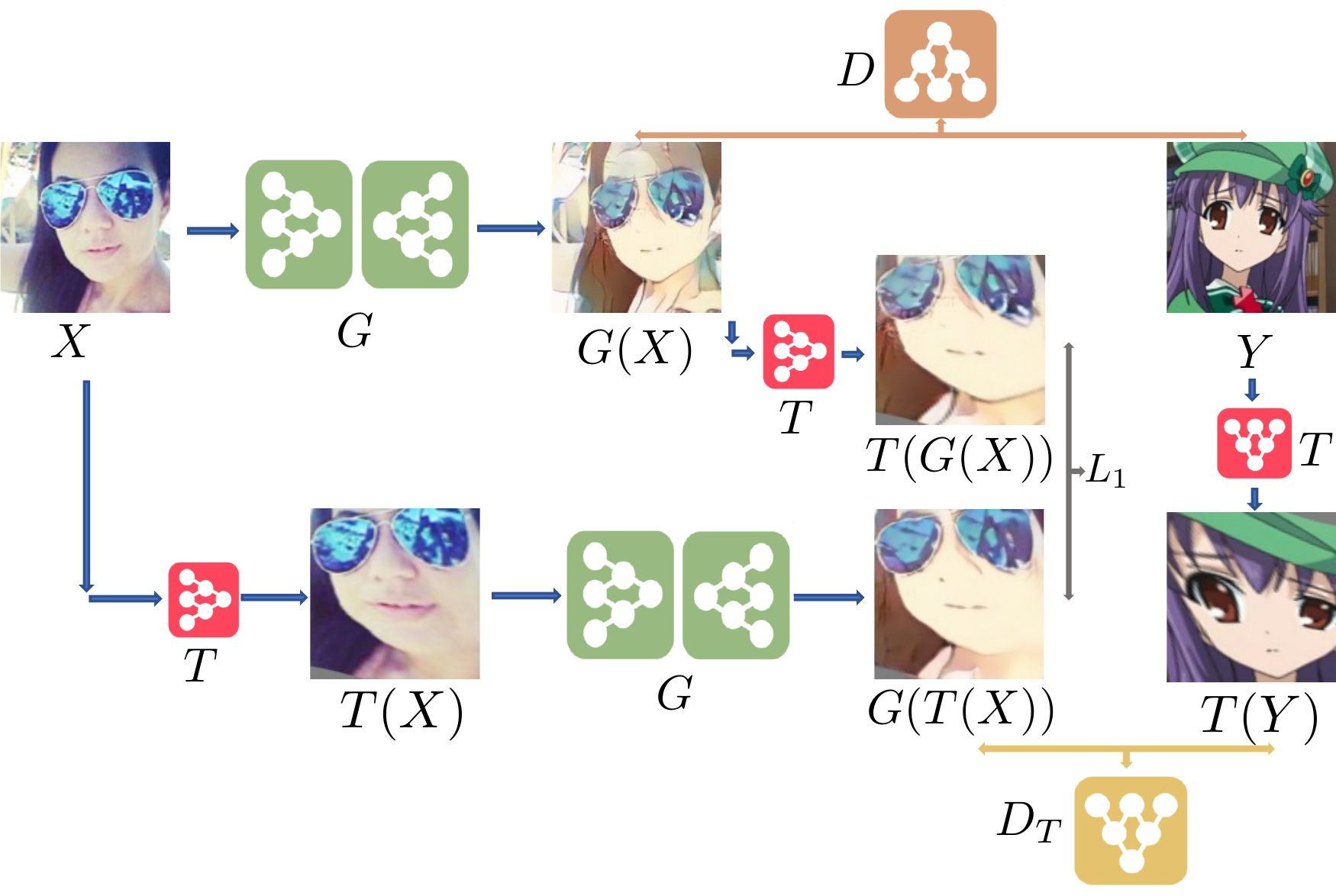} } 
\vspace{4mm}
\subfloat[Maximum Spatial Perturbation Consistency]{
	\label{skematic}
	\includegraphics[width=0.4\textwidth]{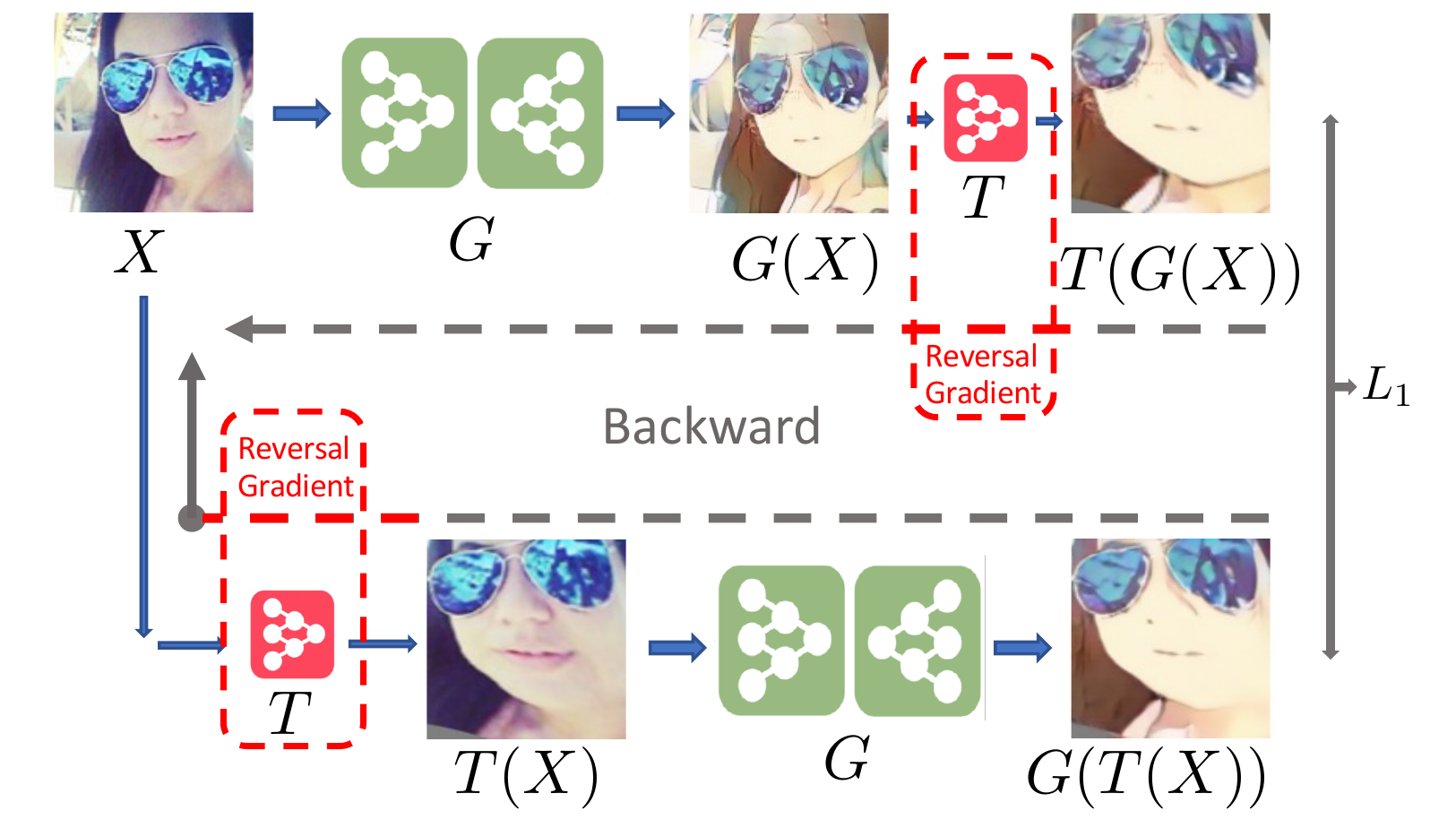} } 
\subfloat[Spatial Alignment of the Transformer $T$.]{
	\label{spatial_alignment}
	\includegraphics[width=0.4\textwidth]{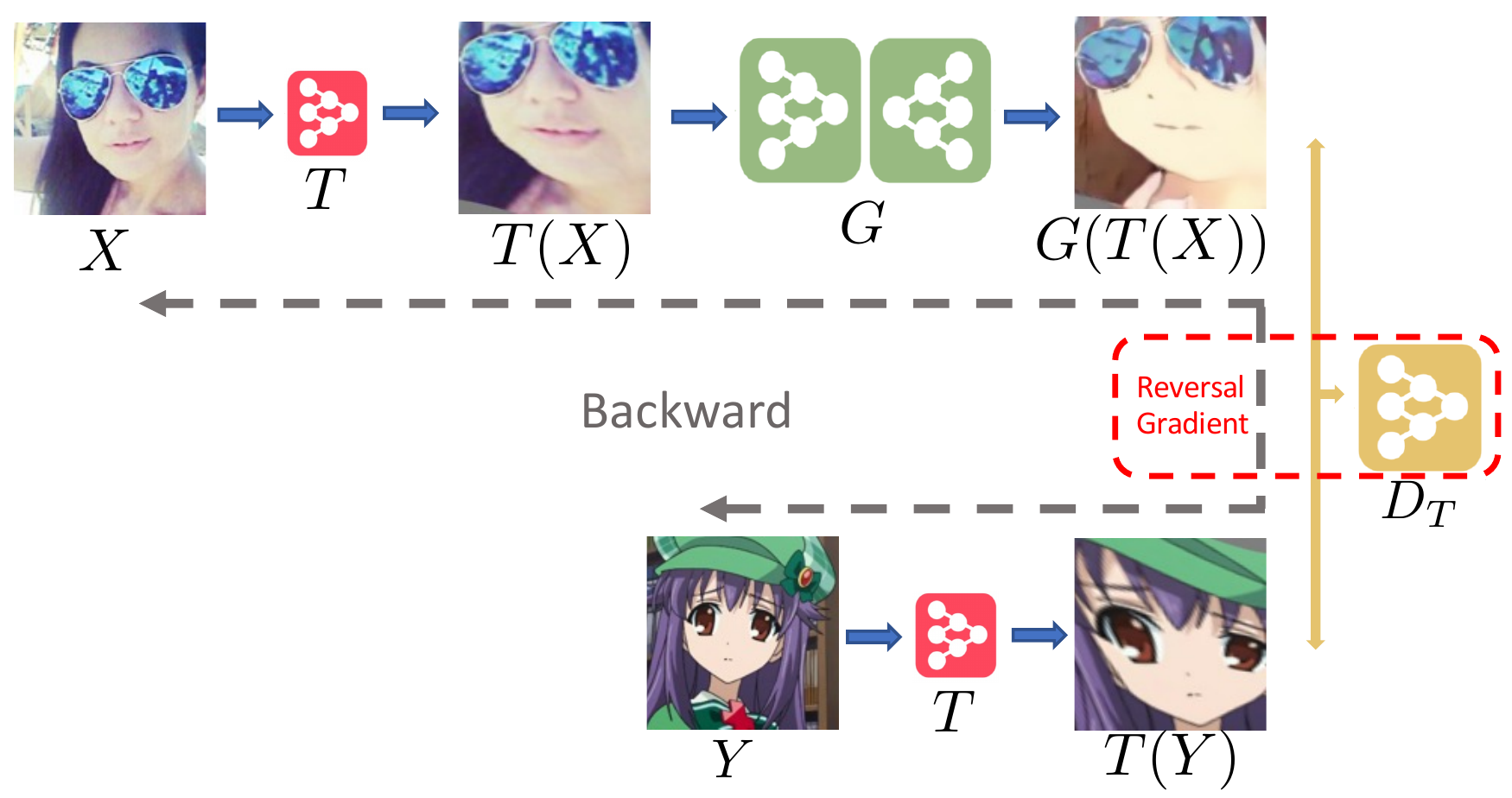} } 
	\vspace{-4mm}
\caption{Illustration of proposed MSPC model, (a)we can summarize our model as three branches of learning 1) $X\to G \to D \leftarrow Y$; 2) $X\to G \to T \to L_1 \leftarrow G \leftarrow T \leftarrow X$; 3) $X\to T \to G \to D_T \leftarrow T \leftarrow Y$. 1), 2), 3) specify the regular adversarial training, maximum spatial perturbation and spatial alignment respectively. To be more specific, we show the adversarial training between $G,T$ in (b) and $G,T,D_T$ in (c) via the forward and backward flow.}
\vspace{-4mm}
\label{unconstrained}
\end{figure*}
In unsupervised I2I, one has access to the unpaired images $\mathcal{X}, \mathcal{Y} \subseteq \mathbf{R}^{C\times H \times W}$, which are from the source and target domains, respectively. The goal is to translate image of $\{x; x\in \mathcal{X}\}$ to $\{y; y\in \mathcal{Y}\}$. Our proposed MSPC has four components and three branches. For the components, we have an image translator $G$, a spatial perturbation function $T$ and two image discriminators $D$ and $D_T$. As the three branches, a) $G$ and $D$ are for regular adversarial training for the image translation; b) $G$ and $T$ compete with each other in the maximum spatial perturbation branch; c) $G$ and $T$ cooperate together to compete with $D_T$ in the spatial alignment branch. The overall architecture of our method is shown in Figure \ref{model}. Below we will explain our method in the order of the branches.
\subsection{Adversarial Constraint on Image Translation}
 A straightforward way of building the translation framework (branch a) is to utilize generative adversarial training \cite{gan}, which forces the translated images to be similar to the target images.
\begin{align}\label{gan}
\min_G\max_D\mathbb{E}_{y\sim P_ Y} \log D(y) + \mathbb{E}_{x \sim P_ X} \log(1-D(G(x))),\nonumber
\end{align}
which is exactly branch a) of our method and has been widely adopted in most I2I approaches \cite{cycle,gcgan,cut}.
\subsection{Maximum Spatial Perturbation Consistency}
In the maximum spatial perturbation branch (branch b), we specify the proposed maximum spatial perturbation consistency (MSPC) for regularizing the unsupervised translation network. Concisely, we propose an adversarial spatial perturbation network $T$ that is to be trained together with the translator $G$. The formulation is as follows:
\begin{align}
    \min_{G}\max_{T}\mathbb{E}_{x\sim P_ X}
    {\left\lVert T(G(x)) - G(T(x)) \right\rVert_1},
\end{align}
where $T$ aims to maximize the $L_1$ distance between the translated image from original input $x$ and the spatial perturbed image $T(x)$, and $G$ learns to minimize the divergence caused by $T$, which is the effect of of spatial perturbation. It is worth noting that $T$ is a parameterized and differentiable network, thanks to \cite{stn}; details will be introduced later. Thus, for each image $x_i$, the learned spatial perturbation $T_i$ is specific to the image. In other words, $T$ generates different spatial perturbations for different images, while in GCGAN, $T$ only represents a fixed spatial transformation. Moreover, our spatial perturbation function $T$ changes as training proceeds. To design the consistency loss, we construct the correspondence between the translated image $G(x_i)$ and the perturbed translated image $G(T_i(x_i))$ via applying the learned $T_i$ on the translated images, which is $T_i(G(x_i))$. A graphic illustration of this branch is given in Figure~\ref{skematic}.

\subsection{Spatial Alignment of the Transformer $T$}\label{section_spatial_alignment}
In branch b), $T$ plays an important role to generate maximum perturbation that tries to confuse $G$ and enable $G$ to be more robust across different I2I tasks. Furthermore, the deforming property of $T$ can help align the spatial distribution in an unsupervised manner between the source images $X$ and the target images $Y$ by scaling, rotating, cropping noisy background, etc. As shown in Figure~\ref{spatial_alignment}, $G$ and $T$ try to force the distribution of $G(T(X))$ to approach the distribution of the transformed target images $T(Y)$ via adversarial training with another discriminator $D_T$. In this process, the target distribution of $P(T(Y))$ is also deformed to be close to the generated distribution, which is different from the regular generative adversarial training with a fixed target distribution. Thus, in the process of c), the adversarial training process can be formulated as the following min-max game,
\begin{align}
\min_{G,T}\max_{D_T}\mathbb{E}_{y\sim P_ Y} \log D(T(y)) + \mathbb{E}_{x \sim P_ X} \log(1-D(G(T(x)))).
\end{align}
\subsection{Differentiable $T$}\label{TSN}
\begin{figure}[H]
\vspace{-3mm}
\centering\includegraphics[width=6cm]{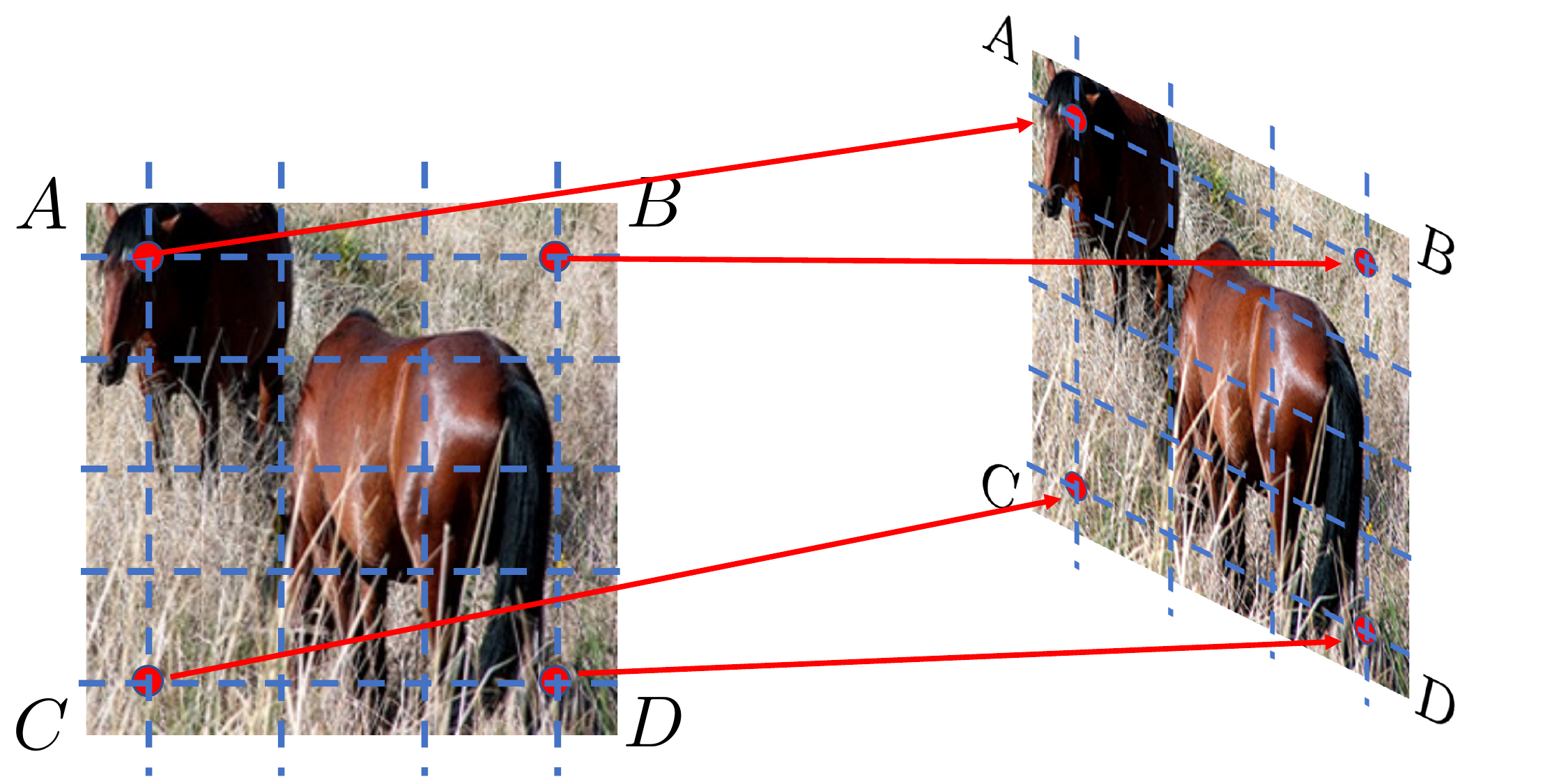}
  \caption{Illustration of spatial transformation network (STN). The network $T$ outputs the coordinates of the deformed grids over the images and then the new images are generated via interpolating in these grids; it is differentiable and can be optimized with stochastic gradient decent.}
  \label{stn}
\end{figure}
All of these functionalities of $T$ in the above sections are based on the nice property that $T$ is differentiable and can be optimized with stochastic gradient decent. According to \cite{stn}, it can be modeled in two steps. In the first step of transforming image, we construct a grid over the image, and the transformation network $T$ outputs the coordinates of the transformed grids. Assuming the image size is $H\times W$, we can simply formulate the process of transformation as
\begin{align}
&\{(p^1_i,p^2_j); i=1,2,3,...,n ,j=1,2,3,...,m\} = T(x), \nonumber\\
&V_{i,j}^c = \sum_{n}^{H} \sum_{m}^W U^c_{nm}k(p^1_i - q^1_m;\Phi_{p^1})k(p^2_j-q^2_n;\Phi_{p^2}), \nonumber \\
&\forall i,m\in[1\ldots H];~\forall j,n\in[1\ldots W]; ~\forall c\in[1\ldots C],
\end{align}
where $(q^1_i,q^2_i)$ represent the coordinate of original grid, $U$ is the pixel value of original image, $c$ is the indicator of image channel, $(p^1_i,p^2_i)$ denotes the new coordinates of transformed grids, $k(;\Phi_{p^1}), k(;\Phi_{p^2})$ represents the kernel of the interpolating image, and we use $V_i$ to denote the transformed pixel value in location $(p^1_i,p^2_i)$. See Figure ~\ref{stn} for an graphic illustration. For the convenience of later formulation, we simply refer to $T(x)$ as the learned transformed image.
\subsection{Constraint on $T$}
\begin{figure}[h]
\centering
\subfloat[Image get distorted heavily without scaling constraint.]{
	\label{scale}
	\includegraphics[width=0.27\textwidth]{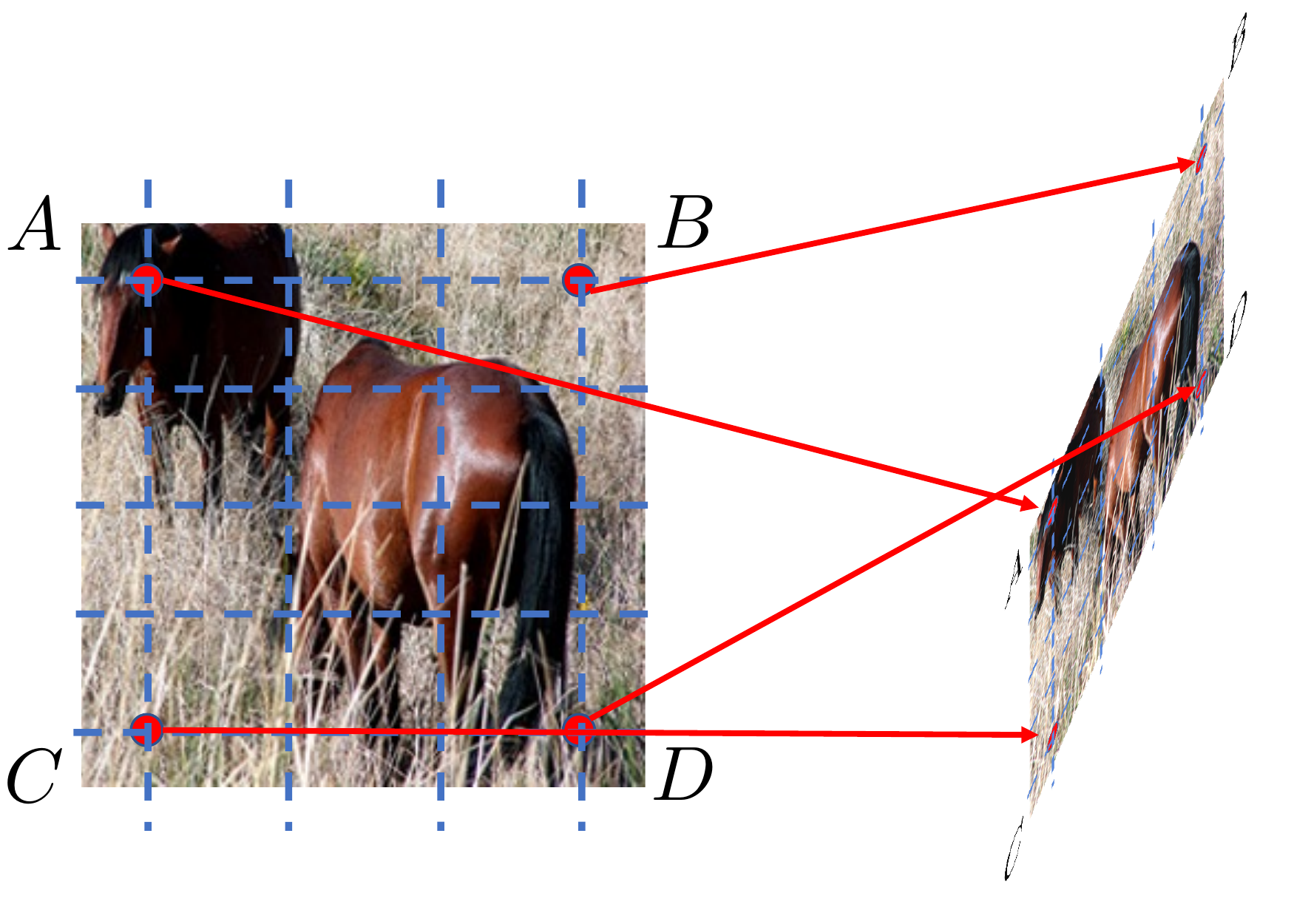} } 

\subfloat[Majority of Image is cropped out.]{
	\label{translation}
	\includegraphics[width=0.27\textwidth]{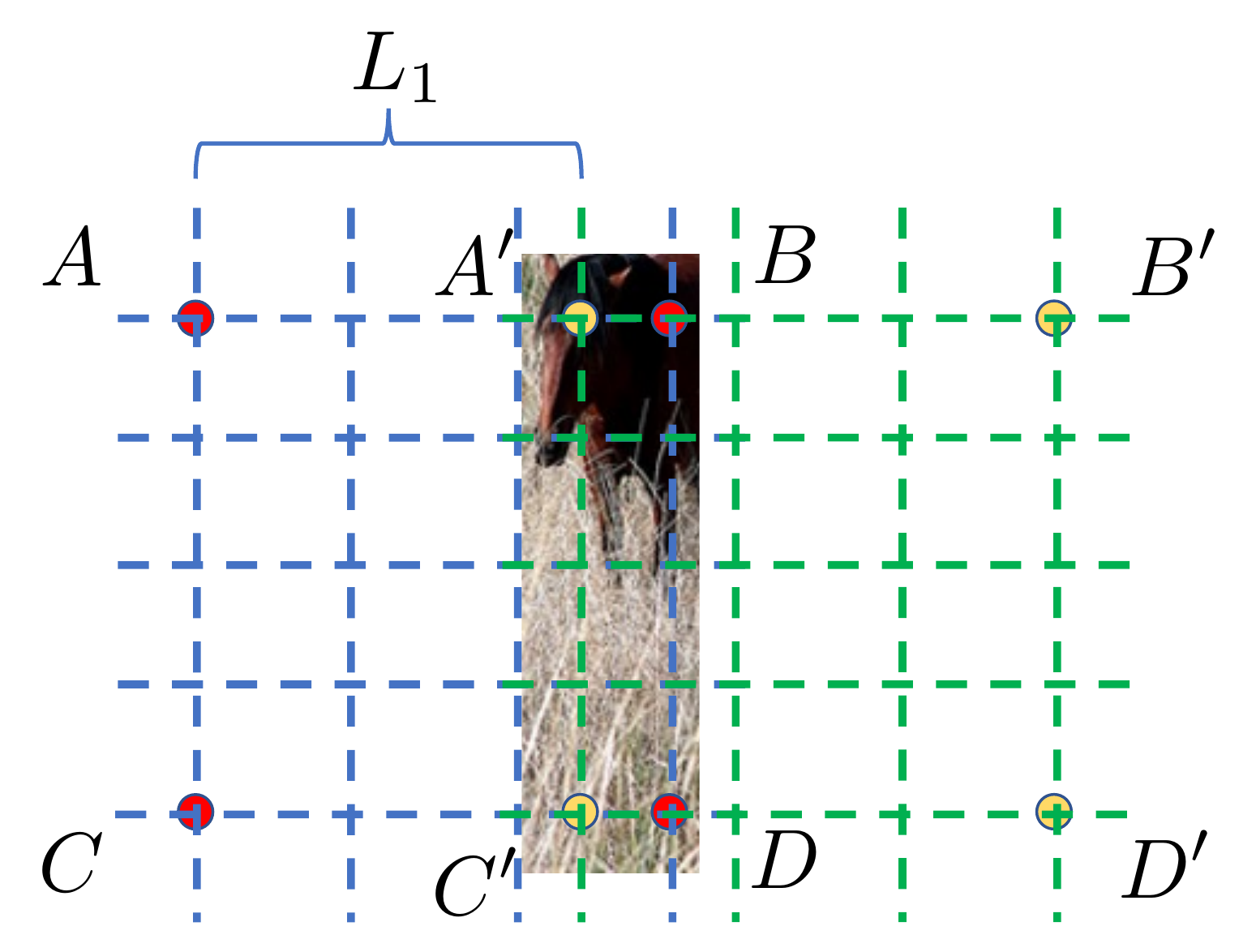} } 
\caption{Illustration of information loss on perturbed images caused by unconstrained $T$.}
\vspace{-4mm}
\label{unconstrained}
\end{figure}
However, without suitable constraints enforced on $T$, $T$ would produce trivial transformation on images, which may worsen the performance of $G$, leading to information loss of images as illustrated in Figure~\ref{unconstrained}. One can naturally come up with an immediate, straightforward way to impose this constraint, for instance, by using
\begin{align}
\label{naive constraint}
     {\left\lVert T(G(x))- G(T(x)) \right\rVert_1} < \epsilon, \, w.r.t. \,\, T.
\end{align}
However, $G$ is a flexible function that can gradually adapt to whatever  transformation learned from $T$, and thus $T$ would still produce transformation beyond the given image distribution. To solve this issue, we directly design a relative scaling constraint of $T$ on the original and transformed coordinates, which is designed to tackle the issue shown in Figure~\ref{scale}. Besides, the major proportion of images would be moved out of the original grids as illustrated in Figure~\ref{translation}, thus we also enforce an absolute constraint on $T$, which restricts the average translation of target coordinates in a reasonable range. According to the property of $T$ as explained in Section~\ref{TSN}, the spatial transformation is based on system of coordinates. Thus, we can directly enforce the relative scaling and the absolute translation constraint on the transformed coordinates, which can be formulated as

\begin{align}\label{constraint of T}
    & \frac{1}{a}<\frac{|p_ip_j|}{|q_iq_j|}<a, \, i\not = j \,\, \& \, -b<\sum^n_{i=1}p_i<b,
\end{align}
where $q_i, p_i$ are the grid coordinates of original and transformed images, respectively, and $a,b$ are constants. The intuition is that, we do not allow the image to be severely distorted beyond a certain scaling and the average translation of coordinates should also be controlled in a reasonable range. The overall formulation of our model can be summarized as follow:

\begin{align}
&\min_{G,T}\max_{D,D_T}\mathbb{E}_{y\sim P_ Y} \log D(y) + \mathbb{E}_{x \sim P_ X} \log(1-D(G(x))) \nonumber \\
&+\mathbb{E}_{y\sim P_ Y} \log D_T(T(y)) + \mathbb{E}_{x \sim P_ X} \log(1-D_T(G(T(x)))),\nonumber \\
&\min_{G}\max_{T}\mathbb{E}_{x\sim P_ X}{\left\lVert T(G(x)), G(T(x)) \right\rVert_1}, \nonumber \\
    & s.t.  \,\, \frac{1}{a}<\frac{|p_ip_j|}{|q_iq_j|}<a, \, i\not = j \,\, \& \, -b<\sum^n_{i=1}p_i<b.
\end{align}

%% file: experiment.tex
\section{Experiment}
\begin{figure*}[h]
\centering\includegraphics[width=13cm]{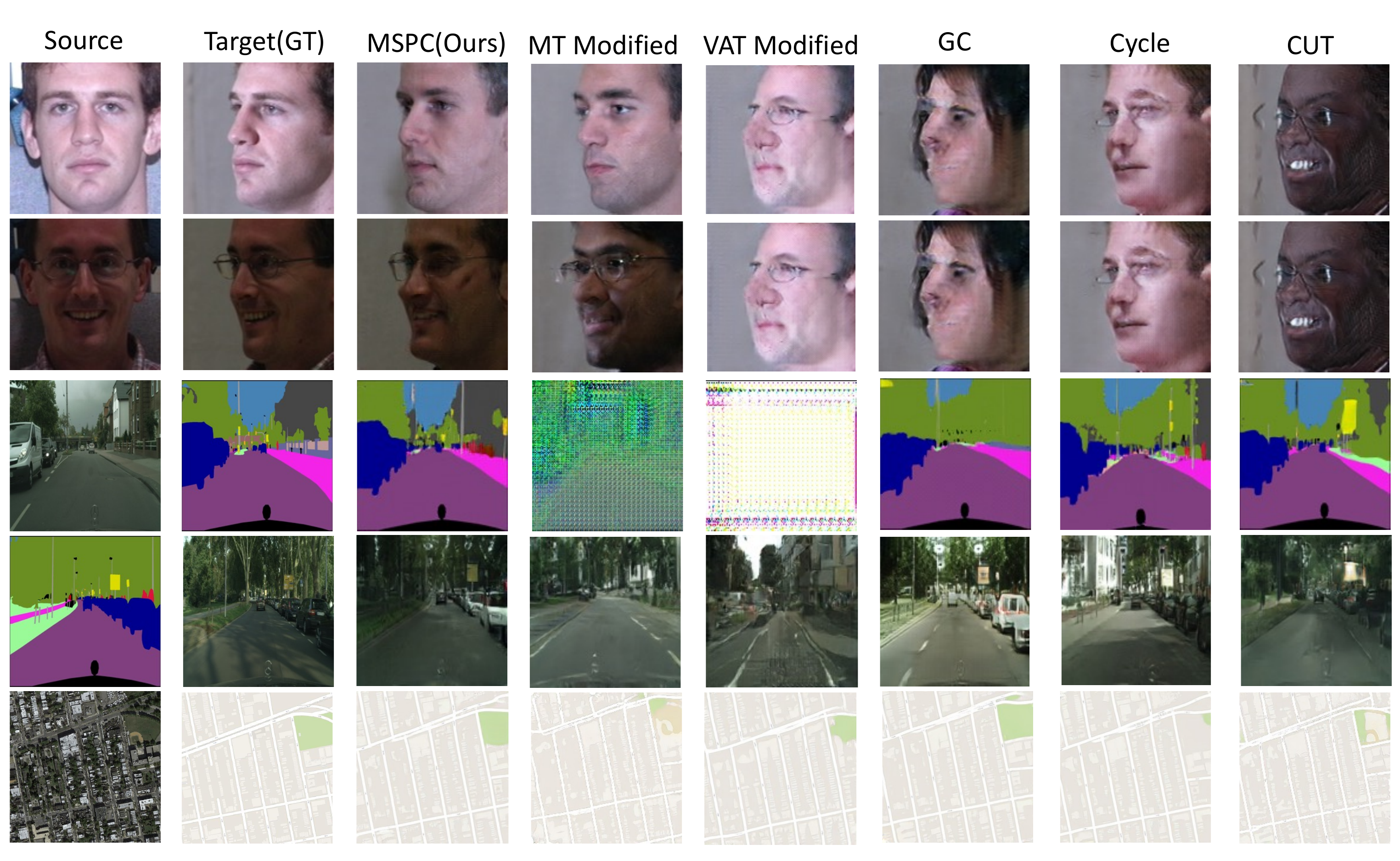}
  \caption{Examples on dataset with paired source and target images, all examples are held out from training dataset. The front face$\to$profile task does not include any paired identity, which is a difficult setting and CycleGAN, GCGAN and CUT cannot be stably trained and collapse in the early training stage. Our model shows a stability across all tasks of image translation.}
  \label{face failure}
\end{figure*}
We conduct  quantitative experiments in different settings on front face$\to$profile, Cityscapes\cite{cityscapes}, Google Map\cite{googlemap}, horse$\to$zebra translations. For  face$\to$profile, we aim to simulate the real-world application, in which we do not have any paired training identities from source to target but evaluate the performance on the held-out front and profile faces with the paired identities. The Cityscapes and Google Map datasets contain paired images in the training datasets, but all the models are trained in an unpaired manner and also tested on paired held-out testing set.s Additionally, we also test the model the on the popular horse$\to$zebra where paired data are not available.
\begin{table*}[t]
  {\small
    \centering
    \resizebox{0.64\linewidth}{!}{
    \begin{tabular}{l @{\hskip 2mm} c@{\hskip 2mm}cc @{\hskip 3mm} c @{\hskip 3mm} c}
    \toprule
    
    \bf Method  & \multicolumn{3}{c}{\bf Cityscapes$\rightarrow$Parsing} & {\bf Front Face$\rightarrow$Profle} & \multicolumn{1}{c}{\bf Horse$\rightarrow$Zebra} \\ %
    \cmidrule(r){2-4} \cmidrule(r){5-5} \cmidrule(r){6-6}
    & {\bf pixAcc}$\uparrow$ & {\bf classAcc}$\uparrow$ & {\bf mAP}$\uparrow$ & {\bf FID}$\downarrow$ & {\bf FID}$\downarrow$ \\
        \midrule
    {CycleGAN}\cite{cycle} & 0.595 & 0.234 & 0.171 & 107.70 & 69.40 \\
        {GCGAN}\cite{gcgan} & 0.563 & 0.195 & 0.143 & 128.31 & 74.89 \\ 
    {CUTGAN} \cite{cut}& 0.587 & 0.225 & 0.166 & 244.50 & 84.26 \\ 
    {MT Modified} & 0.121 & 0.055 & 0.018 & 52.95 & 62.28 \\ 
    {VAT Modified} & 0.484 & 0.100 & 0.064 & 145.54 & 70.21 \\ 
    {MSPC (ours)} & \bf 0.740 & \bf 0.296 & \bf 0.226 & \bf 37.01 & \bf 61.2 \\
    \bottomrule
    \bf Method  & \multicolumn{3}{c}{\bf Parsing$\rightarrow$Cityscapes} & \multicolumn{2}{c}{\bf Aerial Photograph$\rightarrow$Map} \\ %
    \cmidrule(r){2-4} \cmidrule(r){5-6}
    & {\bf pixAcc}$\uparrow$ & {\bf classAcc}$\uparrow$& {\bf mAP} $\uparrow$ & {\bf RMSE} $\downarrow$ & {\bf PixACC} $\uparrow$ \\
        \midrule
    {CycleGAN} \cite{cycle}& 0.508 & 0.184 & 0.117 & \bf 32.70 & \bf 0.265 \\
        {GCGAN} \cite{gcgan}& 0.583 & 0.201 & 0.128 & 33.12 & 0.264 \\ 
    {CUTGAN} \cite{cut}& \bf 0.681 & \bf  0.243 & \bf 0.172 & 35.45 & 0.222 \\ 
    {MT Modified} & 0.455 & 0.145 & 0.086 & 35.43 & 0.216 \\ 
    {VAT Modified} & 0.281 & 0.109 & 0.053 & 63.38 & 0.042 \\ 
    {MSPC (ours)} & 0.612 & 0.214 & 0.156 &  32.97 & \bf 0.265 \\
    \bottomrule
    \end{tabular}
    }
    \caption{\small \textbf{Comparison with baselines} on four dataset with quantitative results, they are conducted on the translation settings of cityscapes$\to$parsing, parsing$\to$cityscapes, front face$\to$profile, horse$\to$zebra and aerial photograph$\to$map respectively. The best scores are bold. Our model shows overall competitive results and robust performance across different settings.
    }\label{allresults}}
\end{table*}

\subsection{Training Configuration}
We unify the model training configuration in this section. We Compare our MSPC, the modified virtual adversarial training (VAT), and the modified mean teacher (MT) models with the recently proposed, popular CycleGAN, GCGAN and CUTGAN, where ``modified" means transferred from semi-supervised framework to I2I. Please refer to the Section 1 in supplementary for the detailed implementation of modified VAT and MT.  We choose the 9-layers of ResNet-Generator with encoder-decoder style \cite{cycle} and the PatchGAN-Discriminator\cite{googlemap} for all of the models. Besides, we choose the Resnet-19 as our $T$ network structure. For all of the model optimization, we set the batch-size to 4 and optimizer to Adam with learning rate $2\times10^{-4}$ and $\beta=[0.5,0.999]$. On all of the dataset, to be fair, we train each model with 200 epoches and we report the performance of the model from the last epoch because of no validation is provided.

Additionally, for our MSPC model, we have three mini-max game between $G,T,D,D_{pert}$. Thus, we separate the model training procedure into two steps, $\{D,D_{pert},T\}-step$ and $G-step$. In each step, we only optimize the corresponding networks and fix others. The size of the spatial transformation grid is $2\times 2$. For all the experiments, we set the maximum scale of perturbation to be $a=\frac{1}{3}, b=3$ and the translation factors to be $c=-0.25, d=0.25$.
\begin{figure*}[t]
\centering\includegraphics[width=13cm]{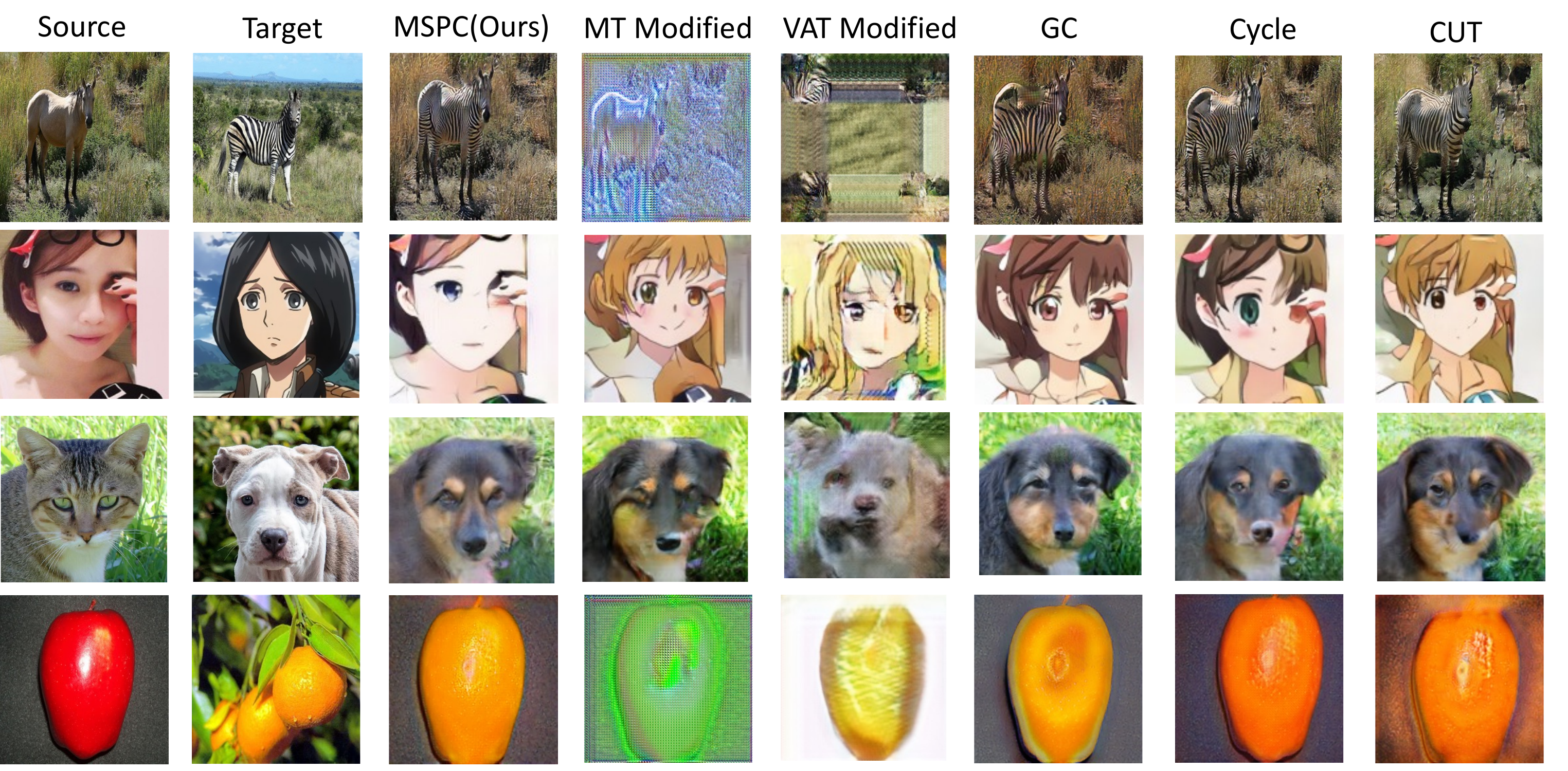}
  \caption{Examples on dataset with unpaired source and target images, all examples are held out from training dataset.}
  \label{qualitative}
\end{figure*}
\subsection{Dataset Configuration and Results}
\paragraph{Front Face$\to$Profile}\label{face_experiment}
In this new dataset, we aim to have an unbiased evaluation metric in real-life applications and explore the possibility of performing the image translation task under a big gap between the source and target domains. To construct such a front face$\to$profile image translation dataset, we sample from CMU Multi-PIE Face \cite{face_angle}, which consists of 250 identities with  different camera angles and the conditions of illumination. We extract two angles of the front and the profile from the dataset and divide them into training and testing sets by different identities. All face images are resized to $128\times 128$.. In the training set, we have 200 identities, 100 in the source and 100 in the target, which do not overlap. For the testing division, we set the source and the target to be paired and calculate the FID score between the translated profile faces and the ground truth of the profile faces. it is worth mentioning that the FID socore is unbiased in this setting due to the paired idenity in the testing set. The lower FID score on the testing set indicates better performance of models.

Quantitative results are shown in Table~\ref{allresults} and  some of the qualitative results are shown in Figure~\ref{face failure}. More qualitative results will be listed in the Section 2 of the supplementary. As we can see from the table and the generated faces, the CycleGan, GCGAN, and CUTGAN failed to stably generate profile from front faces and that our model of the MSPC and the modified MT can generate the faces with high fidelity. Furthermore, except our model, all the remaining models fail to translate front face to the profile while keeping the identity. This illustrates that our model is robust to the large domain gap in the image translation task.
\paragraph{Cityscapes} consists of  city scene images and the mask-level annotation, which can be used to test the ability of model to discover the correspondence between data and labels. There are 3,975 images with paired segmentation mask, 19 categories, and 1 ignored class. We follow the standard training setting of \cite{cycle,gcgan,cut}: the dataset is separated into the 2,975 and 500 samples for training and testing. The original resolution of the image is $1024\times 2048$. During the training, the images are resized to $128\times128$ for city$\to$parsing direction. For the parsing$\to$image synthesis, we first resize images into $144\times144$ and then randomly crop images to be $128\times128$. In this experiment, we are trying to explore how well the models can discover the semantics without paired labels.

For the evaluation on cityscapes dataset, we follow the same protocol of \cite{cityscapes,fully_seg,cycle}. We report the average pixel accuracy, class accuracy, and the mean IOU with respect to the ground truth. To evaluate the quality of parsing$\to$image synthesis, we utilize the pre-trained FCN \cite{googlemap} to extract the predicted segmentation map.
\paragraph{Aerial photo$\to$Map}
The setting of the dataset is similar to Cityscapes and is obtained from the Google Map\cite{googlemap}. It contains 1096 training images and 1098 testing images. We conduct the translation in direction Aerial photo$\to$Map. The images are resized to $256\times 256$. The RMSE and the pixel accuracy are reported across different models.
\paragraph{Horse$\to$Zebra}
For the Horse$\to$Zebra translation scenario, we test if the model is capable of handling the case of real-life applications. The dataset is re-sampled from ImageNet \cite{imagenet}. The source dataset includes 939 horse images and the target includes 1177 zebra images from the wild. The images are resized to $256\times 256$. Because there are no paired images in the testing set, the FID score is biased and reported for reference only.

Overall, our model gain a competitive performance on all dataset settings and shows a very robust generality. We found that  CUT achieves high scores of semantic segmentation on the Parsing$\rightarrow$Cityscapes task and that CycleGAN has the best results on Aerial photo$\rightarrow$Map. On the remaining datasets, our model always achieves the best results under the same settings. CUT owns the feature of maximizing the mutual information, which can translate images well on a setting without changing much semantic information. The bijective assumption of CycleGAN is suitable for the Map dataset. More qualitative results are shown in Figure~\ref{qualitative}, which are operated on horse$\to$zebra, selfie$\to$anime, cat $\to$ dog, and apple$\to$orange. One can see that the proposed MSPC can preserve the image features well and does not cause unnecessary change of the background, which shows the ability of the spatial alignment of the proposed MSPC.
\begin{figure*}[t]
\centering\includegraphics[width=12cm]{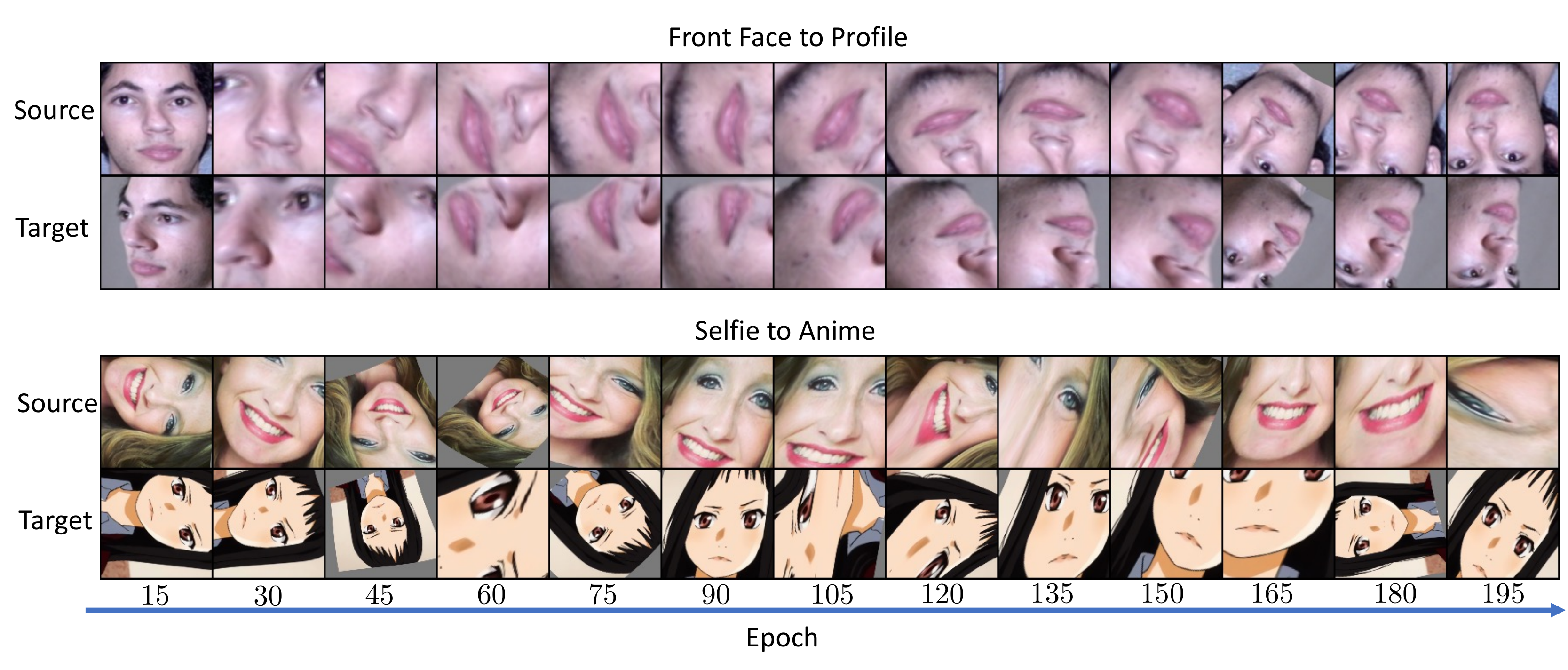}
\vspace{-4mm}
  \caption{Perturbation changes as epoch grows.}
  \label{face_epoch}
\end{figure*}

 \subsection{Ablation Study}
 
 \vspace{-5mm}\paragraph{Effect of scale of perturbation}
  \begin{table}[h]
  \centering
    \small
    \centering
    \resizebox{0.8\linewidth}{!}{
 \begin{tabular}{cccccc}
  \toprule
   \multicolumn{6}{c}{Front Face $\rightarrow$ Profile, changing scaling factor $a$. FID $\downarrow$.} \\
    \midrule
    $a=1$ & $a=2$ &$a=3$ &$a=5$ &$a=8$ & RSP \\
    \midrule
    42.19 & 41.82 & 37.01 & 38.72 & 60.21 & 67.33 \\
  \bottomrule
 \end{tabular}
 }
 \caption{This tables shows the results of the proposed MSPC under different scales of perturbation by changing the scaling factor of $a$ as well as the random spatial perturbation (RSP) for comparison.}\label{ablation perturbation}
\end{table}
 To study the effect of perturbation on model performance, we change the scaling factor of the proposed scaling constraint and conduct the experiments on the front face$\to$profile setting and report the FIDs. To show the effectiveness of maximum perturbation, we also compare with the model of random spatial perturbation (RSP) in Table~\ref{ablation perturbation}, in which the spatial transformation is randomly sampled from the fixed and predefined spatial transformation of rotation, cropping, zoom in, zoom out, stretching, and squeezing. The results in Table~\ref{ablation perturbation} shows that in a certain range of perturbation, more severe perturbation leads to better performance. However, if the perturbation goes beyond image distribution, e.g., images get unreasonably distorted, the performance of MSPC would be cut back. Also the visualization of perturbation without the constraint in Equation~\ref{constraint of T} is shown in the Section 3 of supplementary. Also, we show the dynamic changing of perturbation during training in Figure~\ref{face_epoch}.

  \begin{table}[h]
  \centering
      \small
    \centering
    \resizebox{0.8\linewidth}{!}{
 \begin{tabular}{cccc}
  \toprule
   \multicolumn{4}{c}{Front Face $\rightarrow$ Profile, divergence between distributions. FID $\downarrow$} \\
    \midrule
    $X$, $Y$ & $T(X)$, $T(Y)$ & $G(X)$, $Y$ & $G(T(X))$, $T(Y)$ \\
    \midrule
    112.69 & 65.81 & 37.01 & 30.85 \\
  \bottomrule
 \end{tabular}}
 \caption{This tables quantifies the effect of spatial alignment by transformer $T$. Each row reports the divergence between listed pairs. $X,Y,T(X),T(Y)$ denote the source images, target images, transformed source images by $T$, and transformed target images by $T$. $G(X)$ is the translated images and $G(T(X))$ represents the translated transformed images.}\label{ablation_alignment}
 \vspace{-5mm}
\end{table}
\vspace{-5mm}\paragraph{Effect of Spatial Alignment of $T$}
As we have mentioned in  Section~\ref{section_spatial_alignment}, the spatial perturbation function also plays a role in aligning the image distributions. We conduct an experiment on front face$\to$profile to demonstrate this effect by comparing the FID score between different data pairs. We listed all the controlling pairs in Table~\ref{ablation_alignment}. $(X,Y)$ denotes the divergence between the original source and target images without image translation or spatial transformation. $(T(X),T(Y))$ is the pair of images of spatial transformed source and target images. $G(X)$ and $G(T(X))$ represent the translated images and the translated spatial transformed images. The divergence of pair of $(T(X),T(Y))$ is smaller than $(X,Y)$, because of the effect of spatial alignment by $T$ only. The divergence is further reduced after both the spatial alignment and the image translation, compared to the pair of $(G(X),Y)$ with only image translation. The result clearly shows that the transformer $T$ is capable of alleviating the discrepancy in distribution between the source and the target via the spatial transformation.

%% file: conclusion.tex
\section{Conclusion}
This paper proposes a general regularization method of maximum spatial perturbation consistency (MSPC) to address the limitations of the popular models for image-to-image translation (I2I), including \cite{cycle,gcgan,cut}. We demonstrate 1) that the proposed MSPC is more robust to different applications; 2) that MSPC can help alleviate the spatial discrepancy between domains, such as the discrepancy caused adjusting the object's size and cropping out the noisy background, and further reduce undesired distortions for the translation network. Our method outperforms the state-of-the-art methods on most of of the I2I benchmarks. We also introduce a new benchmark, namely, the front face to profile face dataset, to emphasize the underlying challenges of I2I for real-world applications. We finally perform ablation experiments to investigate the sensitivity of our method to the severity of spatial perturbation and its effectiveness for distribution alignment.
\section{Acknowledge}
This work was partially supported by NIH Award Number 1R01HL141813-01,
NSF 1839332 Tripod+X, SAP SE, and Pennsylvania Department of Health. We
are grateful for the computational resources provided by Pittsburgh SuperComputing grant number TG-ASC170024.
MG is supported by Australian Research Council Project DE210101624. KZ would like to acknowledge the support by the National Institutes of Health (NIH) under Contract R01HL159805, by the NSF-Convergence Accelerator Track-D award \#2134901, and by the United States Air Force under Contract No. FA8650-17-C7715.

%% file: Supplementary.tex
\section{Supplementary}

\subsection{Additional details of MSPC}

\begin{algorithm}
\footnotesize
\caption{\label{algo:optimization} Two steps of optimization of mini-batch MSP for Eq. 7, $\alpha$ denotes the learning rate, $\text{max}(\cdot,\cdot)$ and $\text{min}(\cdot,\cdot)$ represent the hinge loss.}

Choose $M$ samples of $x^{(i)} (i=1,\dots,M)$ and $N$ samples of $y^{(j)} (j=1,\dots,N)$ from $\mathcal{X,Y}$ respectively.

\textbf{Optimizing $D,D_{pert}, T$}
\begin{enumerate}
\item $\mathcal{R}_1={\tiny\frac{1}{N}\sum_{j=1}^{N}\log D(y_j)+\frac{1}{M}\sum_{i=1}^{M}\log(1-}$

${D(G(x_i)))}$
\item $\mathcal{R}_2 = {\tiny\frac{1}{N}\sum_{j=1}^{N} \log D_T(T(y_j))+\frac{1}{M}\sum_{i=1}^{M}}$

${\log(1-D_T(G(T(x_i))))}$
\item $\mathcal{R}_3 = \frac{1}{M}\sum_{i=1}^{M}{\left\lVert T(G(x)), G(T(x)) \right\rVert_1}$
\item $p = T_p(x),\mathcal{R}_4=\frac{1}{M}\sum_{i=1}^{M} [\text{max}(\frac{|p_{jk}p_{jl}|}{|q_{jk}q_{jl}|},a)$

$-\text{min}(\frac{|p_{jk}p_{jl}|}{|q_{jk}q_{jl}|},\frac{1}{a})+|(\sum_k^Kp_{jk})-b|]$
\item $\theta_{D} := \theta_{D} + \alpha\nabla_{\theta_{D}}\mathcal{R}_1,\theta_{D_T} := \theta_{D_T} + \nabla_{\theta_{D_T}}\mathcal{R}_2,$

$\theta_{T} := \theta_{T} - \alpha\nabla_{\theta_{T}}(\mathcal{R}_2-\mathcal{R}_3+\mathcal{R}_4)$
\end{enumerate}

\textbf{Optimizing $G$}
\begin{enumerate}
\item $\mathcal{R}_1={\tiny\frac{1}{N}\sum_{j=1}^{N}\log D(y_j)+\frac{1}{M}\sum_{i=1}^{M}\log(1-}$

${D(G(x_i)))}$
\item $\mathcal{R}_2 = \frac{1}{M}\sum_{i=1}^{M}{\left\lVert T(G(x)), G(T(x)) \right\rVert_1}$
\item $\mathcal{R}_3 = {\tiny\frac{1}{N}\sum_{j=1}^{N} \log D_T(T(y_j))+\frac{1}{M}\sum_{i=1}^{M}}$
\item $\theta_{G} := \theta_{G} - \alpha\nabla_{\theta_{G}}(\mathcal{R}_1+\mathcal{R}_2+\mathcal{R}_3)$
\end{enumerate}
\end{algorithm}
 
 \begin{table}[h]
\renewcommand\thetable{A}
 \caption{This tables shows the results of the proposed MSPC and MSPC without the spatial alignment branch in Fugure 2(c) for comparison. To show the stability, we run each setting for 5 times and calculate the mean and std.}\label{ablation spatial_alignmen}
  \centering
    \small
    \centering
    \resizebox{0.8\linewidth}{!}{
 \begin{tabular}{cc}
  \toprule
   \multicolumn{2}{c}{Front Face $\rightarrow$ Profile. FID $\downarrow$.} \\
    \midrule
    MSPC & MSPC without spatial alignment \\
    \midrule
    $38.61\pm 2.57$ & $53.41\pm 4.83$ \\
  \bottomrule
 \end{tabular}
 }
 \vspace{-3mm}
\end{table}

\subsection{Implementation of modified VAT and MT}

\subsubsection{Modified Virtual Adversarial Training (VAT)}
VAT \cite{vat} introduced the concept of adversarial attack \cite{adversarial_attack} as a consistency regularization in semi-supervised classification. This method learns a maximum adversarial perturbation as a additive , which is on the data-level. To be more specific, it finds an optimal perturbation $\gamma$ on an input sample $x$ under the constraint of $\gamma < \delta$. Letting $\mathcal{R}$ and $f$ denote the estimation of distance between two vectors and the predicted model respectively, we can formulate it as:
	\begin{equation}
		\min_f\max_{\gamma;\|\gamma\|\leqslant \delta}\mathbb{E}_{x\in P_X}\mathcal{R}( f( \theta ,x) ,f( \theta ,x+\gamma )).
		\label{equ:VAT}
	\end{equation}

To apply the VAT, we adapt the semi-supervised framework to the the I2I task. Similar to our proposed MSPC, we introduce another noisy perturbation branch with additional discriminator $D_V$. Then, we can reconstruct the framework as follows,

\begin{align}
&\min_{G}\max_{D,D_T}\mathbb{E}_{y\sim P_ Y} \log D(y) + \mathbb{E}_{x \sim P_ X} \log(1-D(G(x))) \nonumber \\
&+\mathbb{E}_{y\sim P_ Y} \log D_V(y) + \mathbb{E}_{x \sim P_ X} \log(1-D_V(G(x+\gamma))),\nonumber \\
&\min_{G}\max_{\gamma;\|\gamma\|\leqslant \delta}\mathbb{E}_{x\sim P_ X}{\left\lVert G(x), G(x+\gamma) \right\rVert_1}.
\end{align}

Referring to \cite{vat}, the optimal $\hat \gamma$ can be derived from the first-order derivative w.r.t. $\epsilon\gamma$ and $\epsilon$ is a very small positive constant, which is $\hat{\gamma} = \frac{\partial{\left\lVert G(x), G(x+\epsilon\gamma) \right\rVert_1}}{\partial \epsilon\gamma}$. The intuition is that, the direction of maximum perturbation is exactly the same as the current derivative. But VAT is trivious due to that VAT is often unstable when the task is becoming more complex. 

\subsubsection{Modified Mean Teacher (MT)}
MT \cite{meanteacher} is a simple yet non-trivious method, which has been successfully applied in many applications \cite{mt1,mt2,mt3}. It utilizes the exponential moving average (EMA) of the learned model as the teacher reference for correction. The modified MT can be formulated as,

\begin{align}
&\min_{G}\max_{D}\mathbb{E}_{y\sim P_ Y} \log D(y) + \mathbb{E}_{x \sim P_ X} \log(1-D(G(x))) \nonumber \\
&+\mathbb{E}_{x\sim P_ X}{\left\lVert G(x), G_{EMA}(x) \right\rVert_1},
\end{align}
where $G_{EMA}$ is the EMA of $G$ and will not participate in the gradient back-propagation.

For both modified VAT and MT, we use the same networks and training configuration as other models.
\subsection{More Qualitative Results}
In this section, we show additional qualitative results from the held-out testing dataset.
\begin{figure*}[h]
\centering\includegraphics[width=13cm]{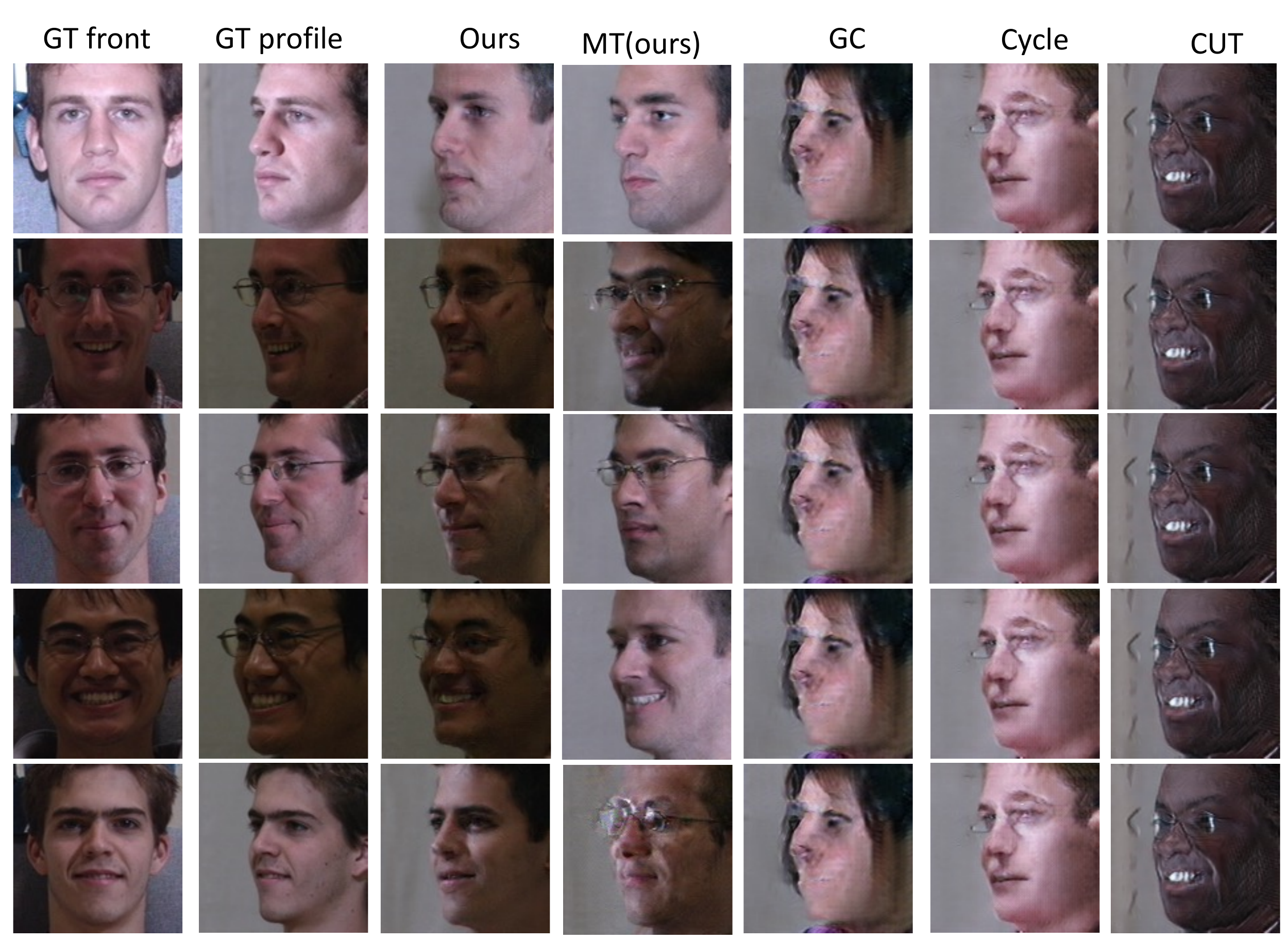}
  \caption{front face2profile.}
  \label{horse2zebra}
  \vspace{-5mm}
\end{figure*}
\begin{figure*}[h]
\centering\includegraphics[width=13cm]{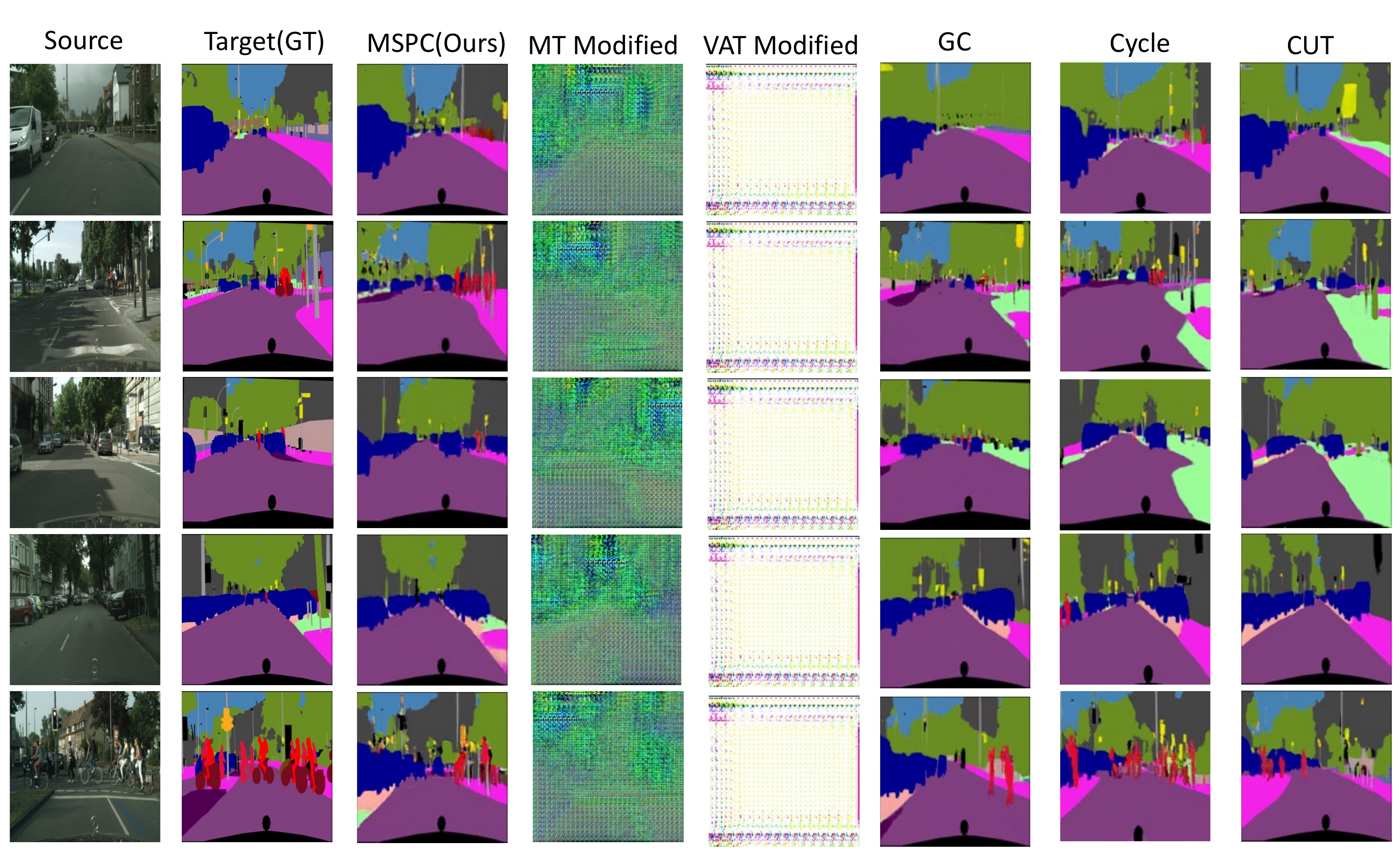}
  \caption{city2parsing.}
  \label{city2parsing}
\end{figure*}
\begin{figure*}[h]
\centering\includegraphics[width=13cm]{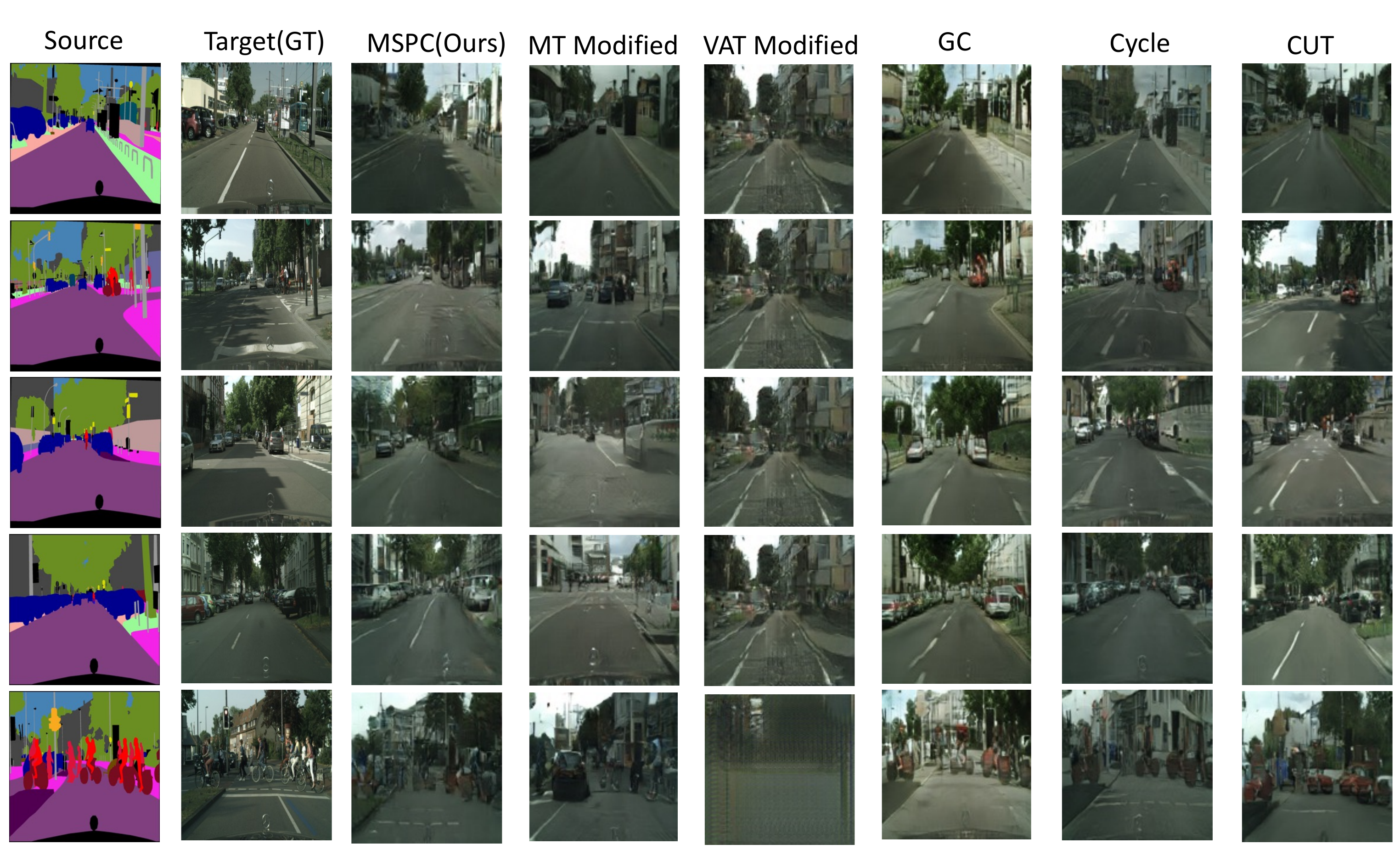}
  \caption{city2parsing.}
  \label{city2parsing}
\end{figure*}
\begin{figure*}[h]
\centering\includegraphics[width=13cm]{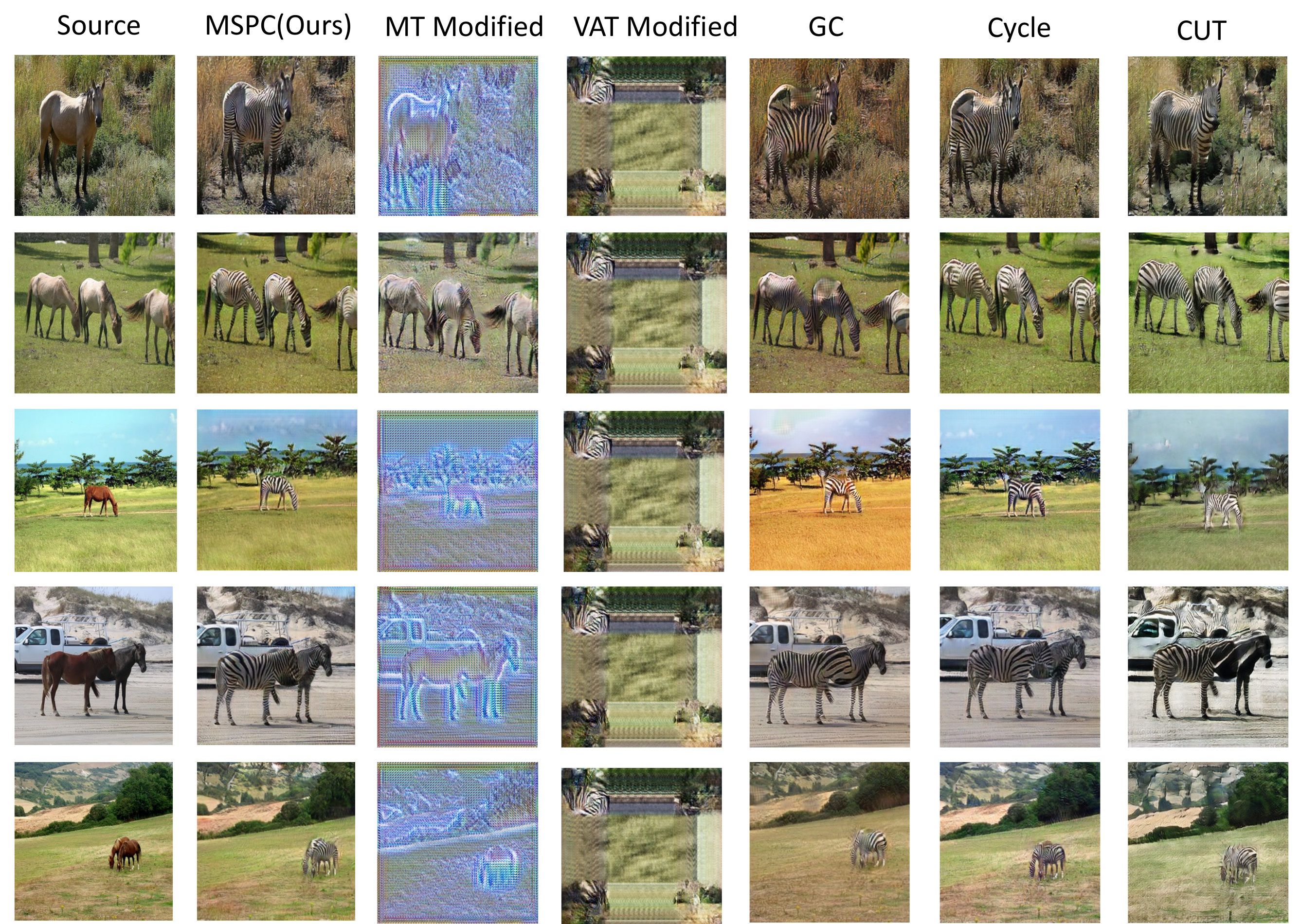}
  \caption{horse2zebra.}
  \label{horse2zebra}
\end{figure*}
\begin{figure*}[h]
\centering\includegraphics[width=13cm]{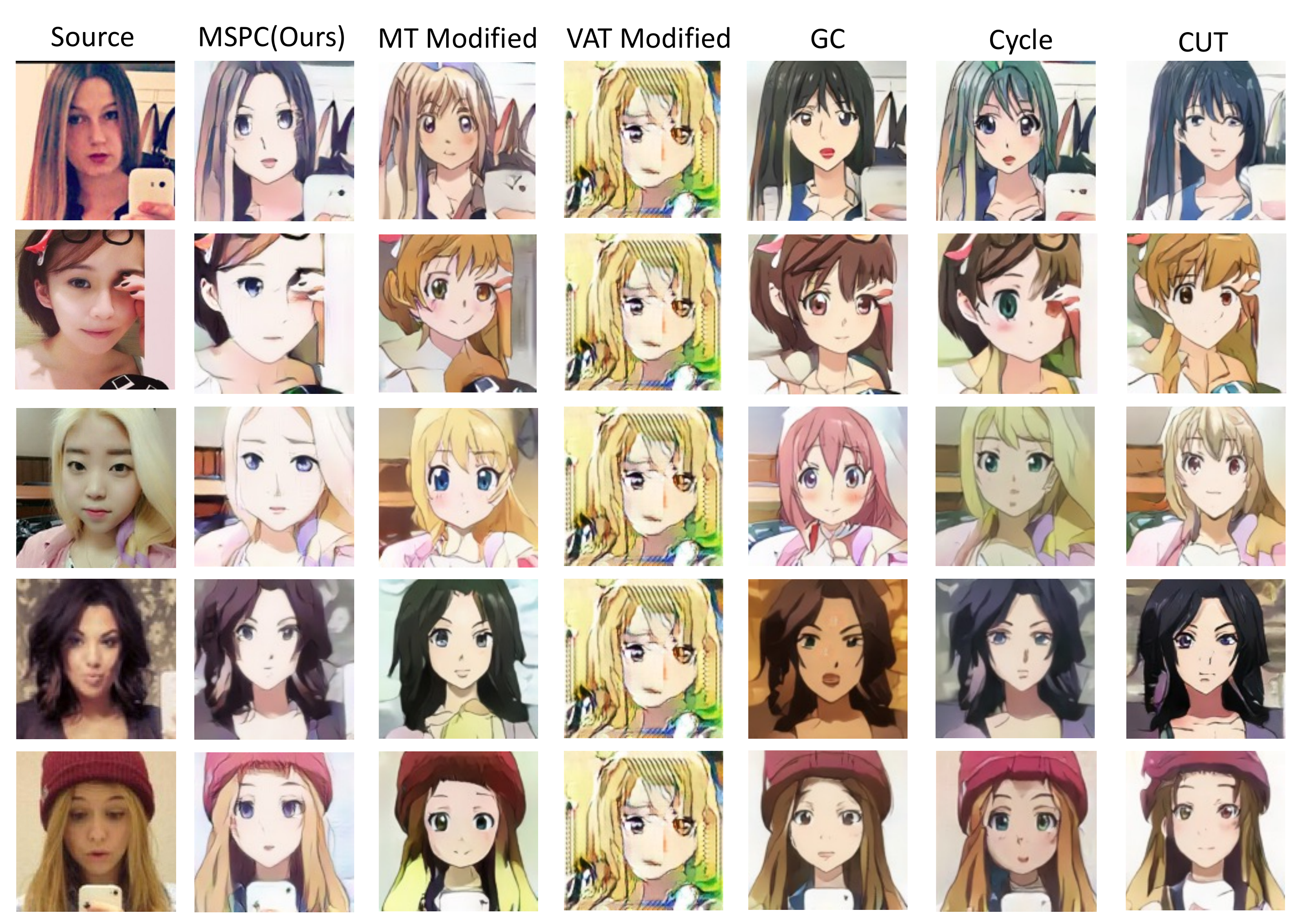}
  \caption{selfie2anime.}
  \label{selfie2aime}
\end{figure*}
\begin{figure*}[h]
\centering\includegraphics[width=13cm]{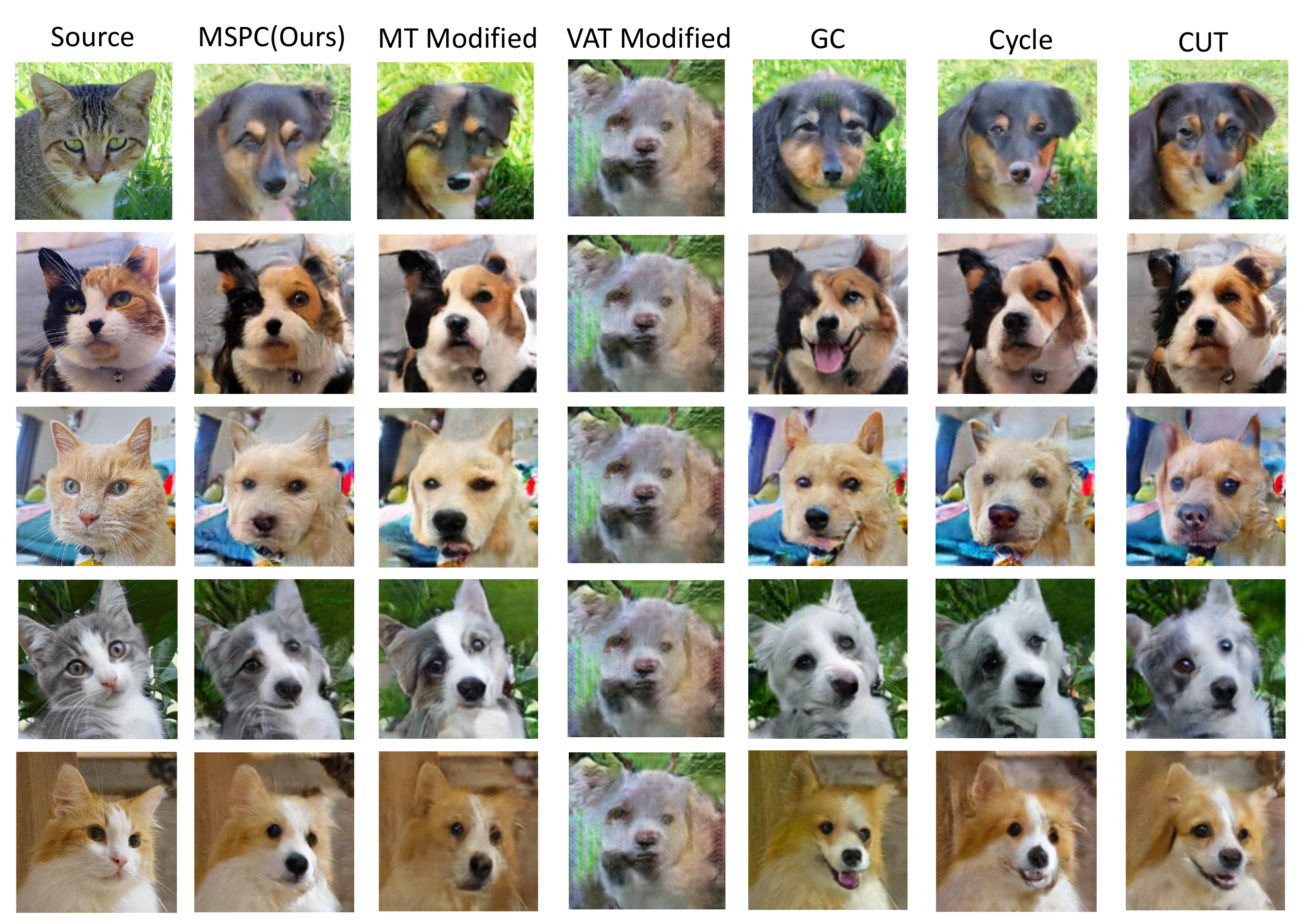}
  \caption{cat2dog.}
  \label{cat2dog}
\end{figure*}
\begin{figure*}[h]
\centering\includegraphics[width=13cm]{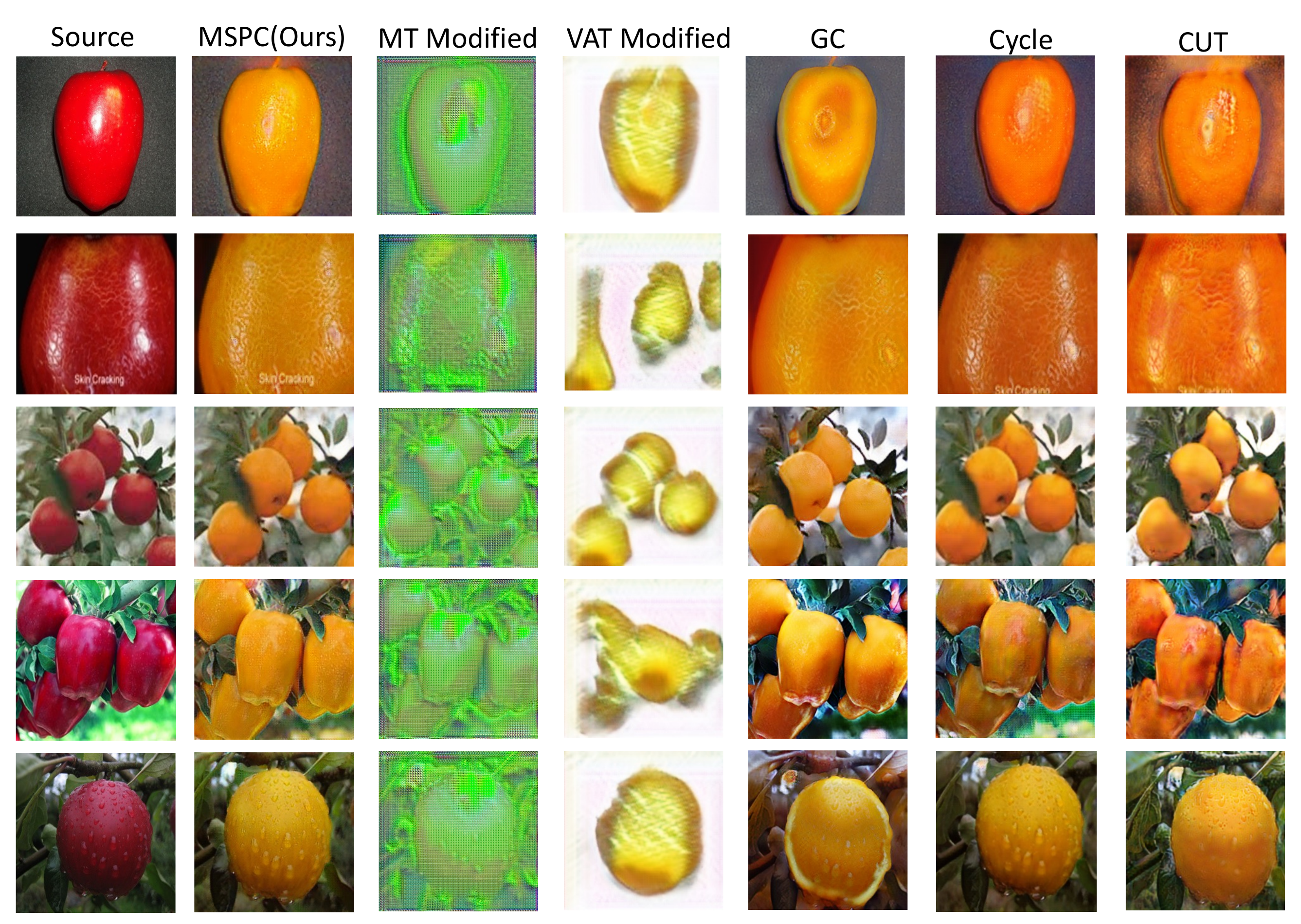}
  \caption{cat2dog.}
  \label{cat2dog}
\end{figure*}

\onecolumn
\section{Visualization of Transformer $T$ without constraints}

In this section, we visualize the effect of the spatial transformer $T$ on both the source and target images. As we can see in below figure, the spatial transformation generates perturbed images without keeping the information of images without the constraint on $T$. If there is much information lost on images, the $T$ will hurt the performance of the I2I.

\begin{figure*}[h]
\centering\includegraphics[width=18cm]{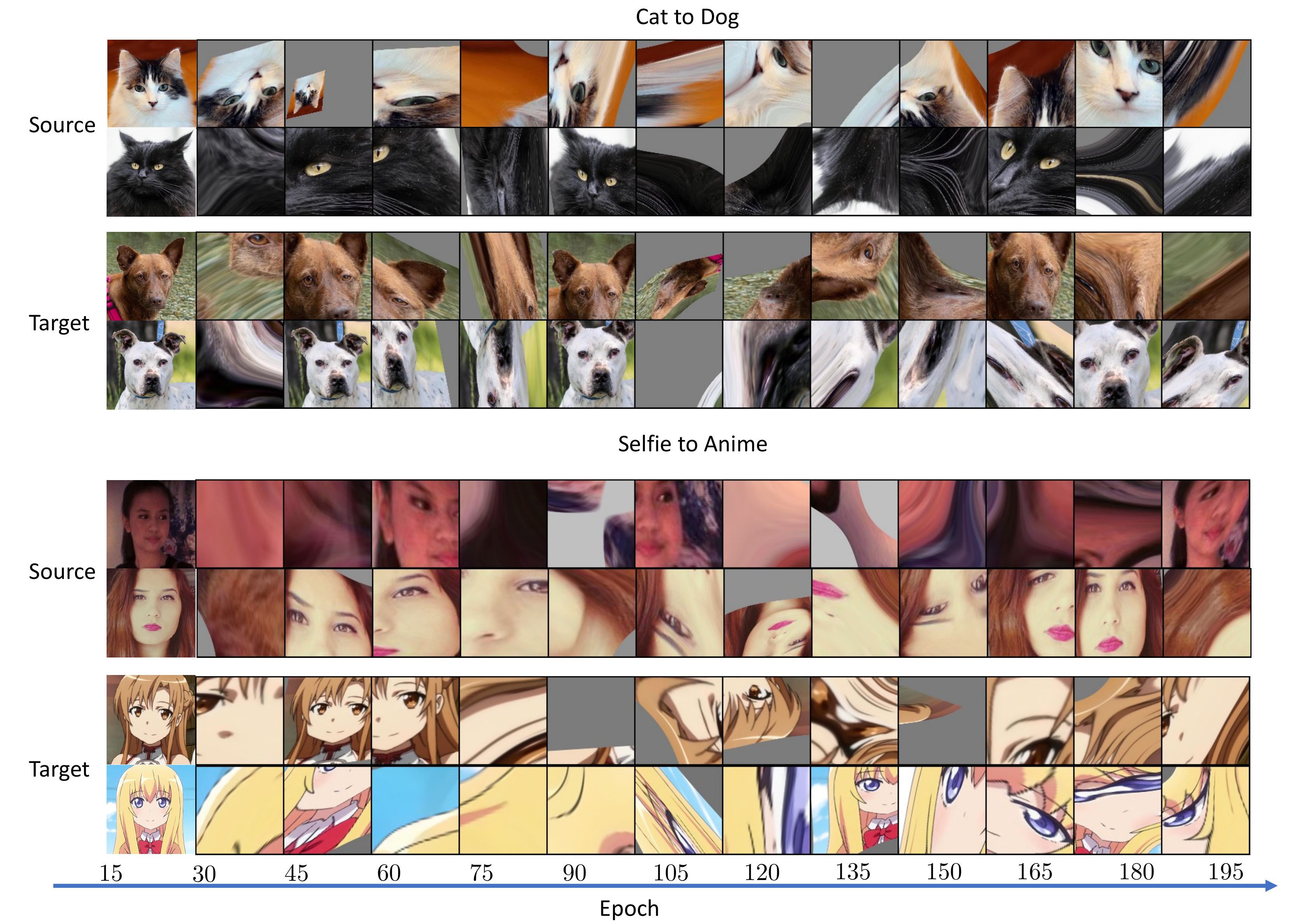}
  \caption{Perturbation changes as epoch grows. In this figure, we do not add the constraint to the $T$.}
  \label{cat2dog}
\end{figure*}